\documentclass[acmlarge]{acmart}
\AtBeginDocument{%
  }

\setcopyright{cc}
\setcctype{by}
\acmJournal{IMWUT}
\acmYear{2026} \acmVolume{10} \acmNumber{2} \acmArticle{33}
\acmMonth{6} \acmDOI{10.1145/3810218}
\usepackage{newfloat}
\usepackage{listings}
\usepackage{booktabs}
\usepackage{multirow}
\usepackage{amsmath}
\usepackage{pifont}
\usepackage{subcaption}
\usepackage{xcolor}
\usepackage[table]{xcolor}
\providecommand{\revision}[1]{#1}
\providecommand{\minorrevision}[1]{#1}

\providecommand{\revisionrow}{}
\begin{document}

\title{\raisebox{-0.5ex}{\includegraphics[height=1.1\baselineskip]{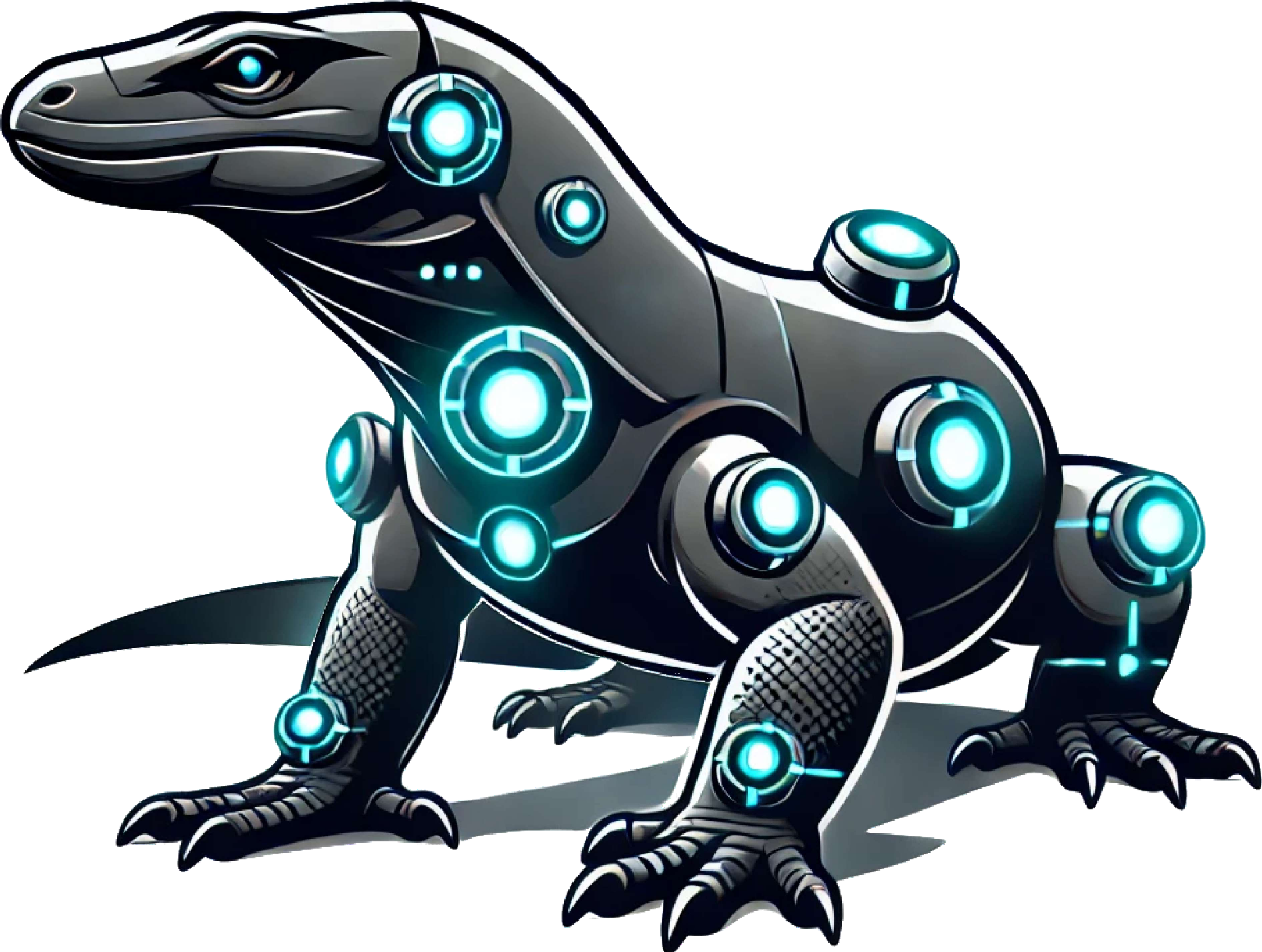}} COMODO: Cross-Modal Video-to-IMU Distillation for Efficient Egocentric Human Activity Recognition}

\author{Baiyu Chen}
\orcid{0009-0000-8617-9635}
\affiliation{%
  \institution{The University of New South Wales}
  \city{Sydney}
  \country{Australia}}
\email{breeze.chen@unsw.edu.au}

\author{Wilson Wongso}
\orcid{0000-0003-0896-1941}
\affiliation{%
  \institution{The University of New South Wales}
  \city{Sydney}
  \country{Australia}}
\email{w.wongso@unsw.edu.au}

\author{Zechen Li}
\orcid{0009-0001-6900-833X}
\affiliation{%
  \institution{The University of New South Wales}
  \city{Sydney}
  \country{Australia}}
\email{zechen.li@unsw.edu.au}

\author{Yonchanok Khaokaew}
\orcid{0000-0003-4297-6274}
\affiliation{%
  \institution{King Mongkut's University of Technology North Bangkok}
  \city{Bangkok}
  \country{Thailand}}
\email{y.khaokaew@unsw.edu.au}
\affiliation{%
  \institution{The University of New South Wales}
  \city{Sydney}
  \country{Australia}}
\email{y.khaokaew@unsw.edu.au}

\author{Hao Xue}
\orcid{0000-0003-1700-9215}
\affiliation{%
  \institution{The Hong Kong University of Science and Technology (Guangzhou)}
  \city{Guangzhou}
  \country{China}}
\affiliation{%
  \institution{The University of New South Wales}
  \city{Sydney}
  \country{Australia}}
\email{haoxue@hkust-gz.edu.cn}

\author{Flora D. Salim}
\orcid{0000-0002-1237-1664}
\affiliation{%
  \institution{The University of New South Wales}
  \city{Sydney}
  \country{Australia}}
\email{flora.salim@unsw.edu.au}

\renewcommand{\shortauthors}{Chen et al.}

\begin{abstract}
The goal of creating intelligent, human-centered wearable systems for continuous activity understanding faces a fundamental trade-off: Egocentric video-based models capture rich semantic information and have demonstrated strong performance in human activity recognition (HAR), but their high power consumption, privacy concerns, and dependence on lighting limit their feasibility for continuous on-device recognition. In contrast, inertial measurement unit (IMU) sensors offer an energy-efficient, privacy-preserving alternative, yet lack large-scale annotated datasets, leading to weaker generalization. To bridge this gap, we propose COMODO, a cross-modal self-supervised distillation framework that transfers semantic knowledge from video to IMU without requiring labels. COMODO leverages a pretrained and frozen video encoder to construct a dynamic instance queue to align the feature distributions of video and IMU embeddings. This enables the IMU encoder to inherit rich semantic structure from video while maintaining its efficiency for real-world applications. Experiments on multiple egocentric HAR datasets show that COMODO consistently improves downstream performance, matching or surpassing fully supervised models, and demonstrating strong cross-dataset generalization. Benefiting from its simplicity and flexibility, COMODO is compatible with diverse pretrained video and time-series models, offering the potential to leverage more powerful teacher and student foundation models in future ubiquitous computing research. The code is available at this repository: \url{https://github.com/cruiseresearchgroup/COMODO}.
\end{abstract}

\begin{CCSXML}
<ccs2012>
   <concept>
       <concept_id>10003120.10003138.10003139.10010904</concept_id>
       <concept_desc>Human-centered computing~Ubiquitous computing</concept_desc>
       <concept_significance>500</concept_significance>
       </concept>
   <concept>
       <concept_id>10003120.10003138.10011767</concept_id>
       <concept_desc>Human-centered computing~Empirical studies in ubiquitous and mobile computing</concept_desc>
       <concept_significance>500</concept_significance>
       </concept>
   <concept>
       <concept_id>10002951.10003227.10003236</concept_id>
       <concept_desc>Information systems~Spatial-temporal systems</concept_desc>
       <concept_significance>500</concept_significance>
       </concept>
   <concept>
       <concept_id>10010147.10010178.10010224.10010240</concept_id>
       <concept_desc>Computing methodologies~Computer vision representations</concept_desc>
       <concept_significance>300</concept_significance>
       </concept>
   <concept>
       <concept_id>10010147.10010257.10010258.10010260</concept_id>
       <concept_desc>Computing methodologies~Unsupervised learning</concept_desc>
       <concept_significance>300</concept_significance>
       </concept>
 </ccs2012>
\end{CCSXML}

\ccsdesc[500]{Human-centered computing~Ubiquitous computing}
\ccsdesc[500]{Human-centered computing~Empirical studies in ubiquitous and mobile computing}
\ccsdesc[500]{Information systems~Spatial-temporal systems}
\ccsdesc[300]{Computing methodologies~Computer vision representations}
\ccsdesc[300]{Computing methodologies~Unsupervised learning}

\keywords{Human Activity Recognition, Ubiquitous Computing, Cross-modal Knowledge Distillation, Cross-dataset Generalization, Multi-modal Learning, Self-supervised Learning, Wearable AI, Human-centered Computing}


\maketitle

\section{Introduction}
\label{sec:intro}

\begin{figure}[t]
  \centering
   \includegraphics[width=1\linewidth]{sec/images/motivation.pdf}
   \caption{\textbf{Motivation:} Egocentric videos provide rich semantics but are impractical for continuous on-device recognition, while IMU sensors are lightweight and energy-efficient yet lack large-scale training data. To bridge this gap, we propose cross-modal, self-supervised distillation to enhance IMU representations by leveraging video knowledge.}
   \label{fig:motivation}
\end{figure}

The vision of human-centered computing hinges on creating technology that seamlessly understands and augments human experiences. A cornerstone of this vision is robust Egocentric Human Activity Recognition (HAR), which has profound applications in healthcare, fitness tracking, and human-computer interaction. The increasing availability of smart glasses, headsets, and next-generation wearables, which often integrate multiple sensors like cameras and inertial measurement units (IMUs), provides an opportunity to realize this vision through rich, multimodal sensing.

Recent advances in large-scale video pretraining \cite{tong2022videomae, bertasius2021space, li2023unmasked} have significantly enhanced video-based HAR. The availability of large-scale datasets \cite{carreira2017quo, caba2015activitynet, goyal2017something, grauman2022ego4d, grauman2024ego} has fueled the success of deep learning models in this domain. Although the video modality achieves high accuracy in human activity recognition, it is often impractical for lightweight devices due to high power consumption, privacy concerns ~\cite{10.1145/3613904.3642242, 10.1145/3613904.3642164, 10.1145/3706598.3713391, 10.1145/3432700}, varying lighting conditions, frequent interruptions, etc. Unlike cameras, IMUs operate independently of lighting conditions and visual occlusions while consuming significantly less power, making them ideal for always-on, on-device activity recognition in resource-constrained scenarios. This efficiency and robustness make IMUs a compelling alternative for HAR on lightweight devices. Despite these advantages, its available training data is significantly more limited. Unlike fields such as computer vision and natural language processing, which have seen significant advancements driven by large-scale datasets, human activity recognition using wearable sensors still relies on relatively small datasets, many of which are collected in controlled environments with scripted activities \cite{zhang2012usc, anguita2013public, reiss2012introducing, chavarriaga2013opportunity}. A key challenge is that annotating IMU data is much more labor-intensive and costly than labeling images or text, as it often requires external reference signals and precise temporal alignment \cite{haresamudram2025past}. This lack of diverse real-world data makes it challenging for data-hungry deep learning models to learn robust representations, limiting their generalization ability in practical applications.

Given the limitations of IMU data (see Figure~\ref{fig:motivation}), a promising direction is to leverage the semantic richness of large-scale video training data and pretrained video models while maintaining the efficiency and privacy advantages of IMU sensors. However, there are several challenges to bridge the latent representation from the video domain to IMU domain: (i) \textit{Knowledge transfer.} Designing objectives and optimization to reliably transfer semantic superior video domain knowledge to IMU domain, which has a significant modality and performance gap \cite{gong2023mmg}. (ii) \textit{Cross-modal alignment.} Aligning heterogeneous video and IMU signals with different temporal resolutions and feature spaces~\cite{10.1145/3717608}. (iii) \textit{Generalization.} Ensuring robustness and generalization across various encoder architectures, activity classes, and sensing devices. (iv) \textit{Data efficiency.} Achieving strong IMU representations with minimal labels via self-supervised signals and scalable training.

\begin{table*}[t]
\centering
\setlength{\tabcolsep}{4.5pt}
\begin{tabular}{lccccc}
\toprule
\textbf{Methods} & \textbf{Inf. Modality} & \textbf{Paradigm} & \textbf{Egocentric} & \textbf{Self-supervised} & \textbf{Structure} \\
\midrule
VideoMAE \cite{tong2022videomae} & Video & Masked Rec. & \textcolor{teal}{\checkmark} & \textcolor{teal}{\checkmark} & - \\
TimeSformer \cite{bertasius2021space} & Video & Cross-Entropy & \textcolor{teal}{\checkmark} & \textcolor{red}{\ding{55}} & - \\
\midrule
DeepConvLSTM~\cite{s16010115} & IMU & Cross-Entropy & \textcolor{teal}{\checkmark} & \textcolor{red}{\ding{55}} & - \\
Attend~\cite{abedin2021attend} & IMU & Cross-Entropy & \textcolor{teal}{\checkmark} & \textcolor{red}{\ding{55}} & - \\
\midrule
CrossHAR \cite{hong2024crosshar} & IMU & Masked Rec. + Contrastive & \textcolor{teal}{\checkmark} & \textcolor{teal}{\checkmark} & - \\
IMUGPT 2.0 \cite{leng2024imugpt} & IMU & Synthetic Generation & \textcolor{teal}{\checkmark} & \textcolor{red}{\ding{55}} & - \\
MuJo \cite{fritsch2025mujo} & IMU & Synthetic + Contrastive & \textcolor{red}{\ding{55}} & \textcolor{teal}{\checkmark} & \textcolor{red}{\ding{55}} \\
IMU2CLIP \cite{moon2023imu2clip} & IMU & Contrastive & \textcolor{teal}{\checkmark} & \textcolor{teal}{\checkmark} & \textcolor{red}{\ding{55}} \\
\midrule
COMODO \textbf{(Ours)} & IMU & \textbf{Distribution Distillation} & \textcolor{teal}{\checkmark} & \textcolor{teal}{\checkmark} & \textcolor{teal}{\checkmark} \\
\bottomrule
\end{tabular}
\caption{Comparison of COMODO with representative HAR frameworks. \textbf{Inf. Modality} denotes the modality used for inference. \textbf{Paradigm} refers to the core learning objective, where ``Rec.'' stands for Reconstruction and ``Contrastive'' denotes standard instance-level contrastive learning (i.e., InfoNCE). \textbf{Egocentric} indicates whether the method is compatible with egocentric data. \textbf{Structure} indicates whether the method preserves the semantic structure from the teacher modality. ``-'' denotes not applicable.}
\label{tab:comparison}
\end{table*}

To tackle the challenges above, we present \textit{CrOss-MOdal video-to-imu DistillatiOn} (COMODO) for efficient egocentric HAR, a novel self-supervised learning framework. Inspired by recent advances in self-supervised learning~\cite{he2020momentum}, knowledge distillation~\cite{fang2020seed, limkonchotiwat2022congen}, and multimodal representation learning~\cite{moon2023imu2clip,fritsch2025mujo}, COMODO is designed to effectively align modalities that differ significantly in temporal resolution, signal characteristics, and feature spaces. As summarized in Table~\ref{tab:comparison}, distinct from prior approaches that rely on heavy video inference~\cite{tong2022videomae,bertasius2021space} or instance-wise contrastive alignment~\cite{moon2023imu2clip,fritsch2025mujo} which neglects the teacher's semantic structure, COMODO distills structured knowledge by aligning the similarity distribution between video and IMU embeddings. This is facilitated by a dynamic instance queue that provides a stable and diverse reference distribution, introducing a beneficial temporal prior that stabilizes the cross-modal optimization process and enhances representation learning. \minorrevision{Practically, COMODO decouples training-time paired video supervision from IMU-only deployment, providing a label-free, model-agnostic way to transfer semantic structure from data-rich video to efficient, privacy-preserving IMU sensing.}

\begin{figure}[t]
  \centering
   \includegraphics[width=1\linewidth]{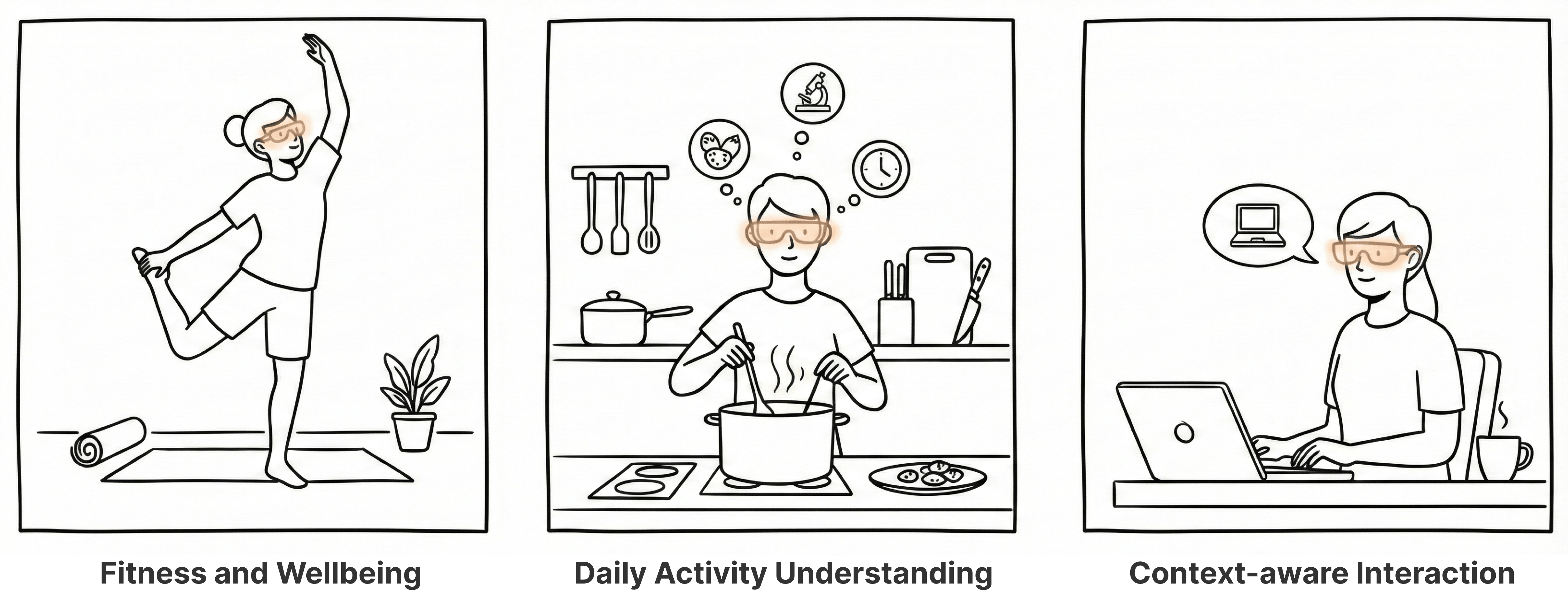}
   \caption{Examples of real-world applications of COMODO-enabled egocentric HAR system. Note: Videos are used only during training to transfer rich semantic knowledge, while lightweight IMU sensors are used at inference for efficient, privacy-preserving operation.}
   \label{fig:application}
\end{figure}

Cross-modal distillation aims to leverage a data-rich modality to improve representations in a data-scarce domain. We show that COMODO enables pretrained time-series encoders to inherit semantic information from video representations, mitigating data scarcity and annotation costs in IMU-based HAR. Evaluated across three egocentric datasets comprising over 70 activity categories, including fine-grained datasets with more than 30 distinct daily activities, COMODO matches or surpasses fully supervised fine-tuned time-series models while retaining IMU efficiency and privacy, suggesting a scalable path to real-world wearable AI (Figure~\ref{fig:application}).

\textbf{Our contributions are summarized as follows:}
\begin{itemize}
    \item \textbf{Self-supervised Cross-modal Knowledge Transfer.} COMODO enables label-free distillation from a stronger (video) to a weaker (IMU) modality using pretrained video and time-series models. Leveraging rich pre-trained knowledge from the video domain to alleviate the challenge of scarce data for wearable IMU sensors.
    \item \textbf{Effective Cross-modal Queuing Mechanism.} We introduce a cross-modal FIFO queue that maintains video embeddings as a stable, diverse reference distribution for IMU distillation, narrowing the modality gap and handling heterogeneity.
    \item \textbf{Model-agnostic Teacher–Student.} COMODO supports diverse pretrained video and time-series models, allowing flexible teacher–student configurations and future integration with stronger foundation models.
    \item \textbf{Cross-dataset Generalization.} To validate real-world applicability, we show that COMODO maintains superior performance on unseen datasets with different classes and devices, outperforming supervised models and highlighting robustness and generalizability for egocentric HAR.
\end{itemize}

% \begin{itemize}
%     \item \textbf{Self-supervised Cross-modal Knowledge Transfer.} COMODO enables label-free distillation from a stronger (video) to a weaker (IMU) modality using pretrained video and time-series models.
%     \item \textbf{Effective Cross-modal Queuing.} We introduce a cross-modal FIFO queue that maintains video embeddings as a stable, diverse reference distribution for IMU distillation, narrowing the modality gap and handling heterogeneity.
%     \item  \textbf{Model-agnostic Teacher–Student.} COMODO supports diverse pretrained video and time-series models, allowing flexible teacher–student pairings and future integration with stronger foundation models.
%     \item  \textbf{Cross-dataset Generalization.} COMODO sustains superior performance on unseen datasets with different classes and devices, outperforming supervised baselines and demonstrating robustness for egocentric HAR.
% \end{itemize}

\section{Related Works}
\label{sec:related_work}

Our work is situated at the intersection of human activity recognition, self-supervised learning, and cross-modal knowledge distillation. To contextualize our contributions, this section reviews prior art from these interconnected domains. We position COMODO by first examining (1) self-supervised learning for HAR, starting from intra-modal sensor-based approaches and extending to cross-modal paradigms. We then discuss (2) the broader field of knowledge distillation, covering its evolution from supervised to self-supervised and cross-modal applications. Finally, we contrast our distillation-based approach with (3) alternative multimodal strategies such as data synthesis and multimodal fusion.

\subsection{Human Activity Recognition}

Human Activity Recognition (HAR) has been explored across both vision and sensor-based modalities, each offering unique advantages. To mitigate the reliance on large-scale labeled sensor datasets, self-supervised learning (SSL) has emerged as a key approach, with contrastive learning effectively capturing generalizable features. Recent works in sensor-based HAR have successfully leveraged unlabeled data through various self-supervised learning strategies. For instance, some approaches utilize a teacher-student framework to distill knowledge from unlabeled data pools \cite{selfhar}, while others creatively treat synchronously captured data from multiple on-body devices as natural transformations for contrastive learning \cite{ColloSSL}. \citet{deldari2022cocoa,deldari2024crossl} introduced contrastive cross-modal learning frameworks for multi-sensor time-series, with \citet{deldari2024crossl} improving upon \citet{deldari2022cocoa} by replacing explicit negative pairs with latent masking. Despite their success, these methods primarily focus on extracting supervisory signals from within the sensor modality itself. \citet{miao2024spatial} proposed a self-supervised framework designed for multi-device HAR that learns robust sensor embeddings via spatial-temporal masking. \citet{hong2024crosshar} introduced hierarchical self-supervised pretraining to improve cross-dataset generalization. Other works have begun to bridge modalities, seeking to leverage richer semantic sources to improve sensor-based HAR, such as \citet{moon2023imu2clip} align IMU data with vision-language models \cite{radford2021learning}, or projecting IMU embeddings into a video-based semantic space for zero-shot recognition \cite{tong2021zero}. \revision{Most recently, MuJo \cite{fritsch2025mujo} learns a joint feature space using synthetic IMU and extracted poses derived from exocentric videos, which limits its applicability in egocentric scenarios.} In contrast to these, our approach performs cross-modal self-supervised distribution distillation from video to IMU.

\subsection{Knowledge Distillation}

Knowledge distillation traditionally transfers knowledge from large teachers to smaller students. \citet{tian2019crd} proposes contrastive representation distillation to align teacher-student embeddings. Similarly, \citet{tan2023egodistill,ni2022cross} distill rich video representations into IMU models. \citet{radevski2023multimodal} employs multimodal distillation, transferring knowledge from RGB, optical flow, and audio features to a single RGB model, demonstrating that multimodal supervision can improve egocentric HAR. \citet{garcia2021distillation} explore mutual distillation, where modality-specific models collaboratively enhance their representations in a supervised setting. For self-supervised learning, \citet{fang2020seed,limkonchotiwat2022congen} utilize self-supervised teachers to improve smaller models using the similarity distribution knowledge distillation, facilitating the generalization of representation in visual and textual domains. \citet{caron2021emerging} frames self-supervised learning as self-distillation with a momentum teacher. \citet{xue2021multimodal} shows that multimodal students can outperform unimodal teachers by learning richer cross-modal representations. Successful knowledge distillation depends on the alignment of modality-general, task-relevant features \cite{xue2023modality}.

Unlike \citet{tan2023egodistill,ni2022cross}, which require either video input at inference or fine-tuning on the target dataset, our approach performs self-supervised cross-modal distillation, aligning sensor and video representations without labeled data. While prior self-supervised knowledge distillation methods \cite{fang2020seed,limkonchotiwat2022congen,caron2021emerging} focus on single-modal representation learning, our method explicitly transfers knowledge across heterogeneous modalities.

\subsection{Multimodal Learning and Synthetic Data Generation for HAR}

Knowledge transfer between modalities has also been explored using synthetic IMU generation. \citet{kwon2020imutube,leng2024imugpt} synthesized IMU signals from video and language models, demonstrating the potential for large-scale HAR by alleviating the need for extensive real-world IMU data collection. More recently, \citet{li2024sensorllm} aligned motion sensor time-series with large language models for HAR, and \cite{li2025zara} further explored agentic, zero-shot HAR from raw motion time-series. These works focus on data synthesis, language-based or agentic sensor understanding for HAR. Multimodal HAR methods often integrate vision, IMU, and other sensors to improve recognition. \citet{zhang2024masked} applied masked autoencoders for joint representation learning of IMU and video, while \citet{gao2023mmtsa} introduced temporal attention to optimize multimodal feature integration. \citet{gong2023mmg} handled missing modality conditions through a combination of modality dropout training and cross-modal contrastive learning. To address domain adaptation in egocentric action recognition, \citet{hatano2024multimodal} applied multimodal few-shot learning. These works reflect the trend of combining multimodal inputs (video, sensor, and language) to overcome data scarcity and domain shifts. 

Compared to these works, we focus on cross-modal self-supervised distillation, leveraging video embeddings to enhance IMU representations without multimodal fusion at inference. Unlike prior contrastive learning methods that align sensor embeddings via instance-wise contrast, our approach distills structured distributional knowledge from video to IMU, enabling robust IMU-based HAR without requiring large-scale labeled datasets.
\section{Methodology}
\label{sec:method}

\begin{figure*}[!h]
  \centering
   \includegraphics[width=\linewidth]{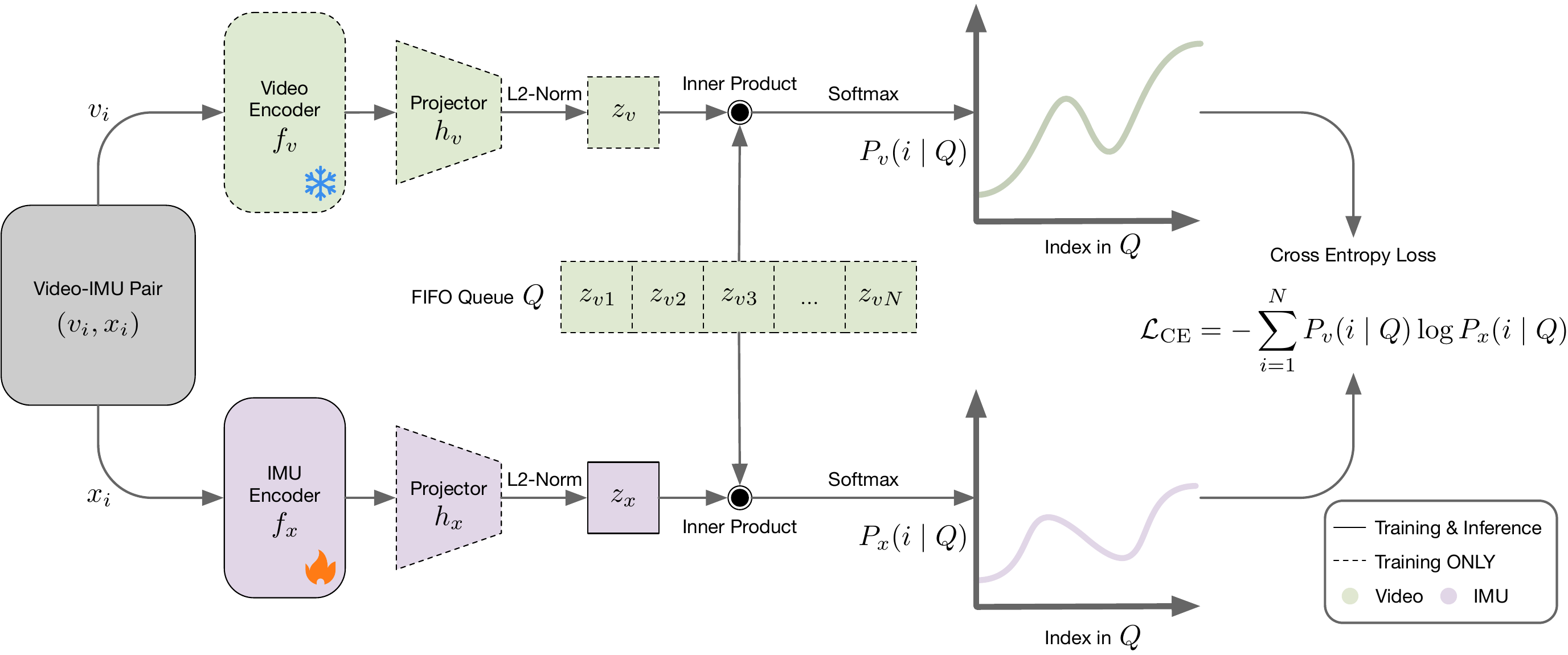}
   \caption{Overview of our cross-modal self-supervised distillation framework. The video encoder is pretrained and kept frozen, while the IMU encoder, initialized from a pretrained time-series model, is trained by minimizing the cross-entropy loss between the similarity distributions of video and IMU embeddings, which are computed based on a continuously updated instance queue.}
   \label{fig:COMODO}
\end{figure*}

As discussed in Section \ref{sec:intro}, while video-based models have demonstrated strong performance in HAR, they come with several practical limitations. On the other hand, IMU sensors are lightweight, power-efficient, and privacy-preserving, making them an attractive choice for real-world applications. However, IMU models lack access to large-scale labeled datasets, making it challenging to learn robust representations. To bridge this gap, we present our COMODO that transfers knowledge from the video modality (which benefits from large-scale pretraining) to the IMU modality without requiring labeled data, as illustrated in Figure \ref{fig:COMODO}.

Contrastive self-supervised learning relies on a sufficiently large number of negative samples to learn discriminative representations effectively. However, maintaining an extremely large batch size is computationally prohibitive. To balance this trade-off, we adopt a FIFO queue approach \cite{he2020momentum}. This design serves two critical purposes in our cross-modal distillation context. First, it provides a computationally efficient mechanism to maintain a large and diverse set of video embeddings, forming a stable reference distribution for the IMU student to learn from. Second, and more subtly, the FIFO structure imposes an implicit temporal prior on the optimization process. Unlike random sampling from a static bank, gradual updating of the queue ensures that the target distribution evolves smoothly over training iterations. As we demonstrate in our ablation study (Section \ref{sec:sampling_method}), this leads to a more stable and coherent optimization flow, which is particularly beneficial for bridging the significant gap between heterogeneous modalities.

% Contrastive SSL relies on a sufficiently large number of negative samples to learn discriminative representations effectively. However, maintaining an extremely large batch size is computationally prohibitive. To balance this trade-off, we adopt a FIFO queue approach \cite{he2020momentum}, enabling the model to maintain a large and diverse set of negative samples to learn from a much broader distribution of video embeddings while keeping the computational cost low.

We define our dataset as \(D = \{(v_i, x_i)\}_{i=1}^{|D|}\), where each sample consists of a video window \(v_i\)  and its corresponding IMU window \(x_i\) . The dataset contains \(|D|\)  paired samples collected from egocentric wearable devices. Our objective is to learn an IMU encoder  \(f_x\)  that maps an IMU window \(x_i\)  to a feature representation \(z_x\) , such that the learned representations capture semantic information from the video modality without requiring labeled data.

\subsection{Cross-modal Self-supervised Distillation}
COMODO employs two separate pretrained encoders to process video and IMU data. Given a paired video-IMU window sample \((v_i, x_i)\) from the dataset, we extract the features using:
(1) Video encoder \(f_v\): A frozen pretrained video encoder that maps the video  \(v_i\) to a feature representation \(f_v(v_i)\). The video encoder remains frozen throughout training to leverage large-scale pretrained video representations.
(2) IMU encoder \(f_x\): A pretrained but trainable time-series encoder that maps the IMU signal \( x_i \) to a feature representation \( f_x(x_i) \). Unlike the video encoder, the IMU encoder is continuously updated during training to distill knowledge from the video modality. To address the modality heterogeneity and potential differences in output dimensionality, we append a projection head after each encoder to map features into a shared embedding space: (1) The video projection head \(h_v\) maps \(f_v(v_i)\) to a normalized embedding \(z_v = h_v(f_v(v_i))\). (2) The IMU projection head \(h_x\) maps \(f_x(x_i)\) to a normalized embedding \(z_x = h_x(f_x(x_i))\). To enforce cross-modal distribution soft alignment, both video and IMU embedding are L2-normalized.

The FIFO queue \(Q\) maintains a large pool of video teacher embeddings. Given a new sample, the encoded video embedding \(z_{vN} := z_v\) of the current batch are enqueued \(Q = \{z_{v1}, z_{v2}, \ldots, z_{vN}\}\), the IMU embedding \(z_x\) and video embedding \(z_v\) are then compared against all video embeddings stored in the queue to compute their similarity scores. To compute the similarity score distributions, we apply the softmax function over the inner product between the query embedding and all stored video embeddings:
\begin{equation}
\label{eq:video_prob}
P_v(i | Q) = \frac{\exp(z_v \cdot z_{vi} / \tau^v)}{\sum_{j=1}^{N} \exp(z_v \cdot z_{vj} / \tau^v)},
\end{equation}
\begin{equation}
\label{eq:imu_prob}
P_x(i | Q) = \frac{\exp(z_x \cdot z_{vi} / \tau^x)}{\sum_{j=1}^{N} \exp(z_x \cdot z_{vj} / \tau^x)},
\end{equation}
where \(P_v(i | Q)\) and \(P_x(i | Q)\) represent the similarity score distributions of the video and IMU embeddings, respectively. \( \tau^v \) and \( \tau^x \) are temperature scaling factors for the video and IMU, and \( \left(\cdot\right)\) represents the inner product between the two embeddings.
To encourage the IMU encoder to capture semantic relationships present in the video modality, we align the similarity distributions of the two modalities by minimizing cross-entropy:
\begin{equation}
\label{eq:loss}
\mathcal{L}_{CE} = - \sum_{i=1}^{N} P_v(i | Q) \log P_x(i | Q)
\end{equation}
After computing cross-entropy loss, the oldest batch of encoded videos \(z_{v1}\) is dequeued to maintain a fixed queue size.

In COMODO, the video encoder \(f_v\) is pretrained and kept frozen. Freezing \(f_v\) ensures that the video embeddings stored in the FIFO queue \(Q\) remain stable throughout training, providing a consistent reference for the IMU encoder to align with. Given that all embeddings are L2-normalized, the similarity score between \(z_v\) and the most recent entry in the queue,  \(z_{vN}\), remains constant at 1 before softmax, as \(z_v = z_{vN}\). This ensures that the highest similarity is always assigned to the latest video sample, while other entries in the queue provide a diverse set of negative samples for contrastive learning. By tuning the temperature factor  \(\tau^v\) and \(\tau^x\), we can control the sharpness of the similarity distribution, adjusting the weight assigned to different video embeddings in the queue. The IMU encoder, trained to minimize cross-entropy with respect to the video similarity distribution, progressively aligns its representation with that of the video encoder, allowing it to leverage the rich semantic knowledge from the video modality.

\subsubsection{\revision{Relation to InfoNCE and theoretical motivation.}}
\label{sec:theoretical_diff}
\revision{Given a FIFO queue $Q=\{z_{v1},\ldots,z_{vN}\}$ of frozen video embeddings, we denote by $P_v(i)$ and $P_x(i)$ the similarity distributions defined in Eq.~\ref{eq:video_prob} and Eq.~\ref{eq:imu_prob}, respectively (with the same queue $Q$).
Our objective in Eq.~\ref{eq:loss} can be rewritten as
\begin{equation}
\mathcal{L}_{CE} = -\sum_{i=1}^{N} P_v(i)\log P_x(i)
\;=\; H(P_v) + \mathrm{KL}(P_v\|P_x),
\end{equation}
where $H(P_v)$ is the entropy of the video teacher distribution and is constant with respect to the parameters of $f_x$ (and $h_x$) since $f_v$ is frozen.
Therefore, COMODO performs \emph{similarity distribution alignment} by minimizing $\mathrm{KL}(P_v\|P_x)$, i.e., distilling the structure of video teacher's similarity space rather than enforcing only instance-wise alignment.

\textbf{InfoNCE as a special case.}
As discussed above, the most recent entry in the FIFO queue corresponds to the current teacher embedding (i.e., $z_{vN}=z_v$), yielding the maximal similarity before softmax.
When $\tau^v \rightarrow 0$, the teacher similarity distribution $P_v(i)$ becomes increasingly peaked and smoothly approaches a one-hot target with $P_v(N)\rightarrow 1$ and $P_v(i\neq N)\rightarrow 0$.
In this extreme case, $\mathcal{L}_{CE}$ reduces to the ``hard'' contrastive objective
\begin{equation}
\mathcal{L}_{NCE} \;=\; -\log\frac{\exp(z_x\cdot z_v/\tau^x)}{\sum_{j=1}^{N}\exp(z_x\cdot z_{vj}/\tau^x)},
\end{equation}
which is the standard InfoNCE-style cross-modal contrastive learning.

\textbf{Why the soft distribution helps in video-to-IMU distillation.}
The gradient of $\mathcal{L}_{CE}$ with respect to the IMU embedding $z_x$ is

\begin{equation}
\label{eq:gradient_analysis}
\frac{\partial \mathcal{L}_{CE}}{\partial z_x}
=
\frac{1}{\tau^x}\left(
\sum_{i=1}^{N} P_x(i)\,z_{vi}
-
\sum_{i=1}^{N} P_v(i)\,z_{vi}
\right),
\end{equation}
which encourages $z_x$ to match the video teacher's similarity distribution over the queue, thereby preserving the structure of the video teacher's similarity space.
In contrast, InfoNCE corresponds to replacing $\sum_{i=1}^{N} P_v(i)\,z_{vi}$ by the single positive $z_v$, (i.e., it treats all non-positives as equally dissimilar and pushes them away uniformly).
For heterogeneous modalities (video vs.\ IMU), many ``negatives'' can be semantically close under the video teacher, (i.e., $P_v(i)>0$ for several $i\neq N$).
In this case, the ``hard'' InfoNCE target (one-hot on the positive) enforces a gradient that ignores these teacher-indicated neighbors, whereas COMODO provides a soft, distribution-level target and updates $z_x$ to fit the teacher's similarity distribution via $\sum_{i=1}^{N} P_v(i)\,z_{vi}$.
Consequently, COMODO provides a distribution-level supervisory signal that preserves the structure of the video teacher's similarity space rather than collapsing it into a single positive instance, which is well aligned with cross-modal distillation under semantic overlap among queue entries. In our experiments (Section \ref{sec:ablation_distillation}), we empirically verify this advantage by comparing COMODO against InfoNCE baseline and L2 alignment.}

\subsection{Inference}
\label{sec:inference}
Once training is complete, only the IMU encoder \( f_x \) is used for inference, while the video encoder \( f_v \), projector networks \( h_v, h_x \), and the FIFO queue \( Q \) are discarded. Given an IMU window \( x \), the trained IMU encoder acts as a feature extractor, producing an embedding \( z_x \). Following prior work on unsupervised representation learning \cite{10.5555/3692070.3692712, yue2022ts2vec,franceschi2019unsupervised}, we evaluate the quality of the learned representations by training a Support Vector Machine (SVM) with an RBF kernel on the extracted features from the training subset, and then using it to predict the classes of the test subset. In this way, our representation learning distillation framework remains unsupervised, with class labels only used to train the SVM on the extracted features.
\section{Experiments}
\label{sec:result}

\subsection{Datasets}
To evaluate the effectiveness of COMODO, we used 3 publicly available egocentric HAR datasets: Ego4D ~\cite{grauman2022ego4d}, EgoExo4D ~\cite{grauman2024ego}, and MMEA ~\cite{10184468}, each with distinct sensor configurations and sampling rates. To ensure consistency and comparability, we apply dataset-specific preprocessing strategies, which we detail below. The statistics of the datasets are summarized in Table \ref{tab:dataset-statistics}.

\subsubsection{Dataset Activity Labels}
\label{sec:labels}
We list below the activity categories available in each dataset.

\paragraph{Ego4D}
The Ego4D dataset contains 31 activity categories after filtering classes with fewer than ten samples: \textit{cleaning / laundry, crafting/knitting/sewing/drawing/painting, cooking, "jobs related to construction/renovation company (director of work, tiler, plumber, electrician, handyman, etc)", household management - caring for kids, farmer, carpenter, scooter mechanic, reading books, walking on street, bike mechanic, playing games / video games, working out at home, working at desk, gardening, playing board games, cycling / jogging, watching tv, practicing a musical instrument, baker, playing with pets, eating, "car - commuting, road trip", potting plants (indoor), playing cards, fixing something in the home, car/scooter washing, bike, working out outside, biology experiments, indoor navigation (walking)}.

\paragraph{EgoExo4D}
EgoExo4D provides 8 coarse activity categories: \textit{Rock Climbing, Cooking, Health, Bike Repair, Basketball, Music, Dance, Soccer}.

\paragraph{MMEA}
The MMEA dataset includes 32 fine-grained daily activities:  \textit{upstairs, downstairs, drinking, fall, reading, sweep-floor, cut-fruits, mop-floor, writing, wipe-table, wash-hand, standing, play-phone, type-PC, eating, cooking, pick-up-phone, drop-trash, fold-colthes, walking, play-card, brush-teeth, wash-dish, moving-sth, type-phone, chat, open-close-door, ride-bike, sit-stand, take-drop-sth, shopping, watch-TV}.

\begin{table}[H]
\centering
\begin{tabular}{lllccc}
\toprule
\textbf{Dataset} & \multicolumn{2}{l}{\textbf{Statistics}} & \textbf{Tra.} & \textbf{Val.} & \textbf{Tst.} \\
\midrule
\multirow{4}{*}{Ego4D} & \multicolumn{2}{l}{\# Media files} & 1110 & 162 & 316 \\
& \multicolumn{2}{l}{Total Media Durations} & 352.2h & 51.9h & 95.7h \\
& \multicolumn{2}{l}{\# IMU$\leftrightarrow$Video Pairs} & 253K & 37K & 68K \\
& \multicolumn{2}{l}{\# Activities} & 31 & 31 & 31 \\
\midrule
\multirow{4}{*}{EgoExo4D} & \multicolumn{2}{l}{\# Media files} & 3495 & 501 & 998 \\
& \multicolumn{2}{l}{Total Media Durations} & 153.8h & 22.7h & 42h \\
& \multicolumn{2}{l}{\# IMU$\leftrightarrow$Video Pairs} & 109K & 16K & 29K \\
& \multicolumn{2}{l}{\# Activities} & 8 & 8 & 8 \\
\midrule
\multirow{4}{*}{MMEA} & \multicolumn{2}{l}{\# Media files} & 4553 & 653 & 1316 \\
& \multicolumn{2}{l}{Total Media Durations} & 21.2h & 3h & 6.2h \\
& \multicolumn{2}{l}{\# IMU$\leftrightarrow$Video Pairs} & 4553 & 653 & 1316 \\
& \multicolumn{2}{l}{\# Activities} & 32 & 32 & 32 \\
\bottomrule
\end{tabular}
\caption{Ego4D / EgoExo4D / MMEA Statistics.}
\label{tab:dataset-statistics}
\end{table}

\subsubsection{Preprocessing}

\textbf{Ego4D} and \textbf{EgoExo4D} are preprocessed following the same strategy as IMU2CLIP~\cite{moon2023imu2clip} to ensure consistency. First, we exclude video samples that do not contain IMU recordings. Then, to ensure sufficient class representation, we further remove activity categories with fewer than 10 samples. We divide the remaining data into train, test and validation sets using a 7:2:1 ratio, we also provide the split files in our code repository. The IMU signals are standardized to a uniform sampling rate of 200 Hz, and video frames are resized to 224 \(\times\) 224 pixels and downsampled to 10 FPS. Both the video and IMU data are segmented into non-overlapping 5-second windows. The final segment is discarded if its duration is less than 5 seconds to ensure uniform window length. With a sampling rate of 200 Hz, each IMU segment corresponds to a fixed sequence length of 5 \(\times\) 200 = 1000 time steps. To adapt to different IMU student models, we use model-specific sequence adjustments: MOMENT-small automatically subsamples sequences that exceed its input length limit, while Mantis applies its resize function to reshape all sequences to a fixed length of 512. These preprocessing steps ensure compatibility with diverse IMU models while preserving the temporal and structural integrity of the original sensor data.

\begin{figure}[ht]
  \centering
   \includegraphics[width=0.7\linewidth]{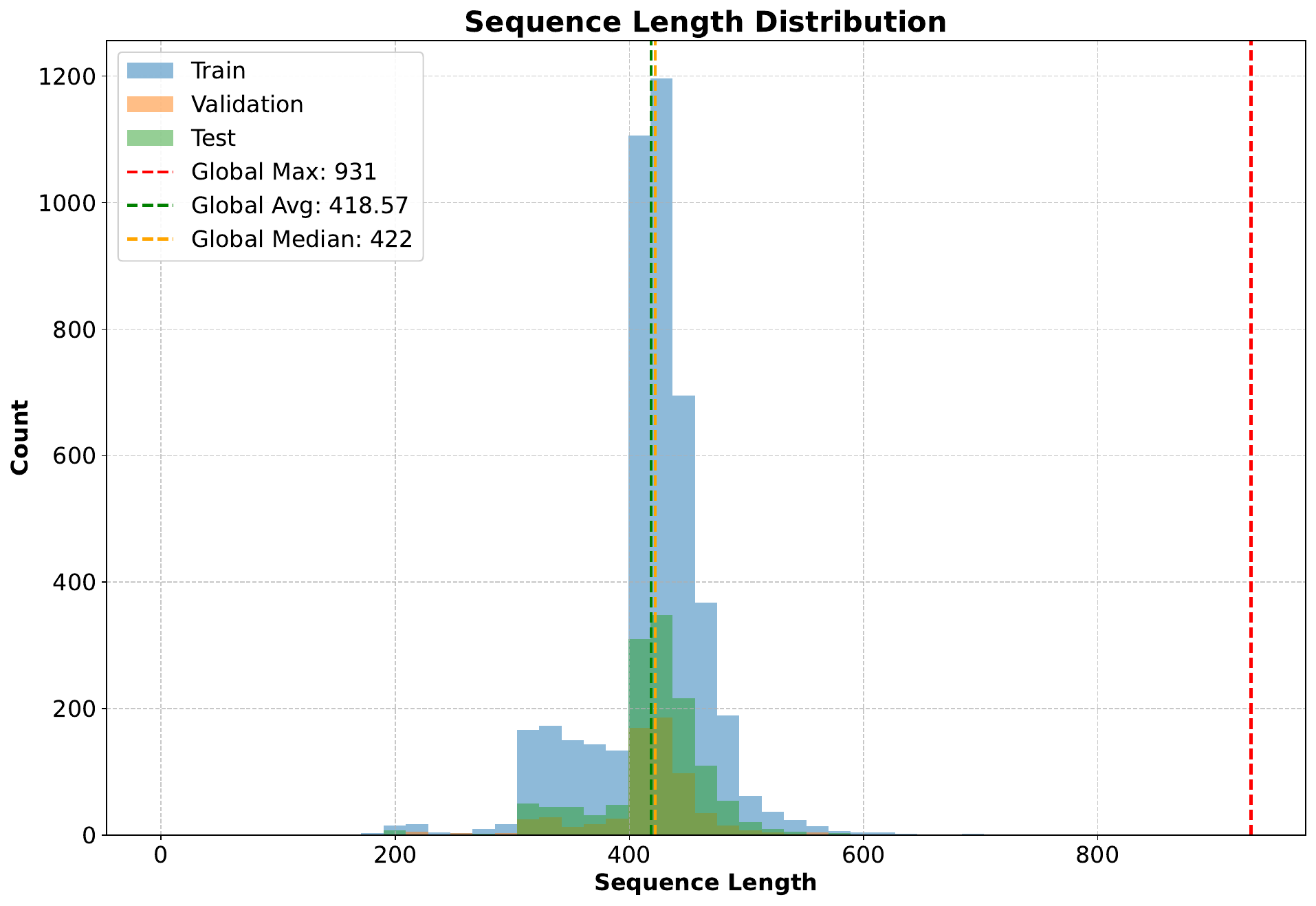}
   \caption{The visualization of distribution of sequence lengths in MMEA dataset.}
   \label{fig:MMEA_dataset}
\end{figure}

\textbf{MMEA} provides raw sensor readings, including accelerometer and gyroscope data, sampled at 25 Hz. This relatively low sampling frequency leads to shorter IMU sequences, with an average sequence length of 418.57 and a median of 422, as visualized in Figure~\ref{fig:MMEA_dataset}. The variance in sequence lengths within MMEA also allow us to evaluate the model's adaptability and robustness to variable-length inputs. To convert raw IMU readings into real-world values, we normalize the data by dividing the accelerometer readings by \(R_{\text{acc}} = 16384\) and the gyroscope readings by \(R_{\text{gyro}}\) = 16.4, resulting in real acceleration values (in g) and angular velocities (in \(^\circ/s\)). The video data in MMEA is resized to 224 \(\times\) 224 pixels, and the frame rate is kept at 25 FPS. We follow the official train-validation-test splits provided by the MMEA dataset.

\begin{table*}[!h]
 % \footnotesize
  \centering
  \setlength{\tabcolsep}{4pt}
    \begin{tabular}{@{}p{2.7cm} c | ccc | ccc | ccc@{}}
        \toprule
        \multirow{2}{*}{\textbf{Methods}} & \multirow{2}{*}{\textbf{Self-supervised}} & \multicolumn{3}{c|}{\textbf{Ego4D}} & \multicolumn{3}{c|}{\textbf{EgoExo4D}} & \multicolumn{3}{c}{\textbf{MMEA}} \\
        \cmidrule(lr){3-5} \cmidrule(lr){6-8} \cmidrule(lr){9-11}
         & & \textbf{A@1} & \textbf{A@3} & \textbf{A@5} & \textbf{A@1} & \textbf{A@3} & \textbf{A@5} & \textbf{A@1} & \textbf{A@3} & \textbf{A@5} \\
        \midrule
        \raggedright DLinear & \textcolor{red}{\ding{55}} & 10.60 & 25.27 & 36.74 & 32.20 & 59.91 & 78.16 & 24.01 & 37.73 & 47.18 \\
        \midrule
        \raggedright Informer & \textcolor{red}{\ding{55}} & 48.89 & 72.75 & 81.75 & 74.60 & 96.28 & 99.17 & 45.81 & 67.68 & 78.81 \\
        \midrule
        \raggedright TimesNet & \textcolor{red}{\ding{55}} & 44.33 & 68.02 & 77.79 & 69.20 & 93.44 & 98.04 & 50.30 & 74.24 & 83.99 \\
        \midrule
        \raggedright DeepConvLSTM & \textcolor{red}{\ding{55}} & 52.85 & 75.70 & 83.49 & 76.51 & 96.89 & 99.37 & 52.58 & 77.25 & 86.44 \\
        \midrule
        \raggedright Attend & \textcolor{red}{\ding{55}} & 58.12 & \underline{78.08} & \underline{84.56} & 81.45 & \underline{98.20} & \textbf{99.66} & 58.66 & 79.87 & 87.47 \\
        \midrule
        \revisionrow
        \raggedright IMUGPT 2.0 (V) & \textcolor{red}{\ding{55}} & 4.25 & 13.44 & 18.77 & 44.76 & 56.80 & 73.93 & 7.15 & 18.48 & 27.77 \\
        \midrule
        \revisionrow
        \raggedright IMUGPT 2.0 (V+R) & \textcolor{red}{\ding{55}} & 46.67 & 71.07 & 80.45 & 74.93 & 96.45 & 99.13 & 17.19 & 35.38 & 47.78 \\
        \midrule
        \revisionrow
        \raggedright CrossHAR & \textcolor{red}{\ding{55}} & 50.96 & 72.66 & 81.53 & 77.14 & 96.95 & 99.31 & 84.97 & 96.55 & 97.74 \\
        \midrule
        \raggedright MOMENT-small & \textcolor{red}{\ding{55}} & 57.59 & 75.91 & 82.94 & 79.26 & 97.04 & 99.33 & 84.27 & 94.76 & 96.88 \\
        \midrule
        \raggedright Mantis & \textcolor{red}{\ding{55}} & \underline{58.36} & 76.98 & 83.76 & \underline{84.22} & 97.95 & 99.41 & \textbf{93.01} & 98.25 & 99.01 \\
        % \midrule
        % \raggedright Chronos-bolt-small & \ding{55} & 46.05 & 69.86 & 79.49 & 70.58 & 94.85 & 98.91 & - & - & - \\
        \midrule \midrule
        \revisionrow
        \raggedright CrossHAR & \textcolor{teal}{\checkmark} & 30.18 & 56.10 & 70.00 & 67.80 & 94.67 & 98.66 & 84.80 & 96.88 & 98.33 \\
        \midrule
        \raggedright MOMENT-small & \textcolor{teal}{\checkmark} & 39.70 & 64.47 & 75.32 & 68.14 & 93.55 & 98.43 & 83.66 & 95.52 & 97.42 \\
        \midrule
        \raggedright Mantis & \textcolor{teal}{\checkmark} & 47.49 & 71.63 & 81.24 & 76.47 & 96.98 & 99.21 & 90.96 & 98.56 & 99.39 \\
        \midrule
        \raggedright IMU2CLIP & \textcolor{teal}{\checkmark} & 54.83 & 76.04 & 84.03 & 79.46 & 97.68 & 99.44 & 91.41 & \underline{98.71} & \underline{99.47} \\
        \midrule
        \raggedright COMODO (\textbf{Ours}) & \textcolor{teal}{\checkmark} & \textbf{59.13} & \textbf{78.79} & \textbf{85.65} & \textbf{84.92} & \textbf{98.28} & \underline{99.59} & \underline{92.48} & \textbf{99.01} & \textbf{99.77} \\
        \bottomrule
    \end{tabular}
  \caption{\textbf{IMU-based Human Activity Recognition} results: We compare COMODO against \revision{14} baselines from both fully supervised and self-supervised approaches. Our method, COMODO, achieves comparable or superior performance to fully supervised models, while consistently outperforming the previous state-of-the-art self-supervised methods, demonstrating its effectiveness in cross-modal transfer learning. The best are indicated in \textbf{bold}, and the second-best are \underline{underlined}. \revision{(V) denotes training with virtual-only IMU data; (V+R) denotes training with virtual + real IMU data.}}
  \label{tab:results}
\end{table*}

% \subsection{Training}
\subsection{Experimental Settings}
\label{sec:training}
\paragraph{Setup.} To evaluate the effectiveness of our proposed framework, we assess the representation quality of the IMU student encoder on two downstream tasks, comparing it against state-of-the-art time-series models, human activity recognition models, and self-supervised learning methods. \revision{Detailed configurations for all baselines, including sequence length and reimplementation specifics, are provided in Appendix~\ref{appendix:baselines}.} We report Accuracy @1, @3, and @5, which reflect both top-1 prediction correctness and the quality of the predicted similarity distribution. In the cross-modal self-supervised distillation stage, we train COMODO for 20 epochs using a batch size of 128 and a learning rate of 3e-4. The temperature hyperparameters in Equations~\ref{eq:video_prob} and~\ref{eq:imu_prob} are set to \(\tau^v = 0.1\) for the video teacher and \(\tau^x = 0.05\) for the IMU student. All experiments are conducted on a single NVIDIA L40S GPU. We fix the random seed to 42 for all experiments to ensure reproducibility.

\paragraph{Video teacher.} By default, we use the pre-trained TimeSformer-base (\#params: 121.4M) ~\cite{bertasius2021space} fine-tuned on Kinetics-400~\cite{carreira2017quo} to leverage the rich knowledge from video modality. We append an MLP projector \(h_v\) at the end of the encoder. The hidden dimension of \(h_v\) is 2048, the output dimension is 128. To ensure efficient training, all video teacher networks remain frozen during distillation. Instead of computing video features on the fly, we precompute and store all video embeddings in a pickle file, allowing direct retrieval during training to reduce computational overhead.

\paragraph{IMU student.} We choose two lightweight time-series foundation models with significantly fewer learnable parameters as the IMU student network: (1) Mantis (\#params: 8.1M) ~\cite{feofanov2025mantis}, the current state-of-the-art classification time-series foundation model which serves as the default choice, and (2) MOMENT-small (\#params: 37.9M) ~\cite{10.5555/3692070.3692712}. To generate a unified IMU representation from multiple sensor channels, we apply concat pooling, where features from each IMU channel are processed separately and then concatenated. Specifically, given an IMU input sequence \(x \in \mathbb{R}^{T \times C} \) with \(C\) sensor channels and \(T\) time steps, we apply a shared time-series model \(f_x\)  independently to each channel: \(z_c = f_x(x_c), \forall c \in \{1, \dots, C\}\), where \(x_c \in \mathbb{R}^{T}\) represents the input sequence for the \(c\)-th IMU channel, and  \(z_c \in \mathbb{R}^{D}\)  is the corresponding feature representation where \(D\) denotes the dimension. The final IMU embedding \(z_x\) is obtained by concatenating all per-channel embeddings: \(z_x = \{z_1, z_2, \dots, z_C\} \in \mathbb{R}^{C \times D}\). This approach ensures that the model captures independent channel-wise representations while maintaining structural consistency in the concatenated feature space. In Section~\ref{sec:pooling-methods}, we present an ablation study showcasing results with different input pooling methods. Similar to the video teacher network, we add one additional MLP projector \(h_x\).

\subsection{IMU-based Human Activity Recognition}
This task aims to classify human activities from IMU sensor data for efficient and reliable recognition on resource-constrained wearables. %Effective IMU-based HAR models should capture motion semantics and generalize robustly across diverse real-world sensor readings. 
We benchmark the classification on three egocentric human activity recognition datasets. Table \ref{tab:results} categorizes methods into supervised and self-supervised settings (Table~\ref{tab:results}):

\paragraph{Supervised methods.} We include methods for time-series and HAR as our supervised baselines: DLinear ~\cite{zeng2023transformers}, Informer ~\cite{zhou2021informer}, TimesNet ~\cite{wu2023timesnet}, DeepConvLSTM~\cite{s16010115}, Attend and Discriminate~\cite{abedin2021attend}, \revision{IMUGPT 2.0~\cite{leng2024imugpt}, CrossHAR~\cite{hong2024crosshar},} MOMENT-small ~\cite{10.5555/3692070.3692712}, and Mantis ~\cite{feofanov2025mantis}. These models are fine-tuned on labeled IMU data in this task.

\begin{figure}[h]
  \centering
   \includegraphics[width=0.7\linewidth]{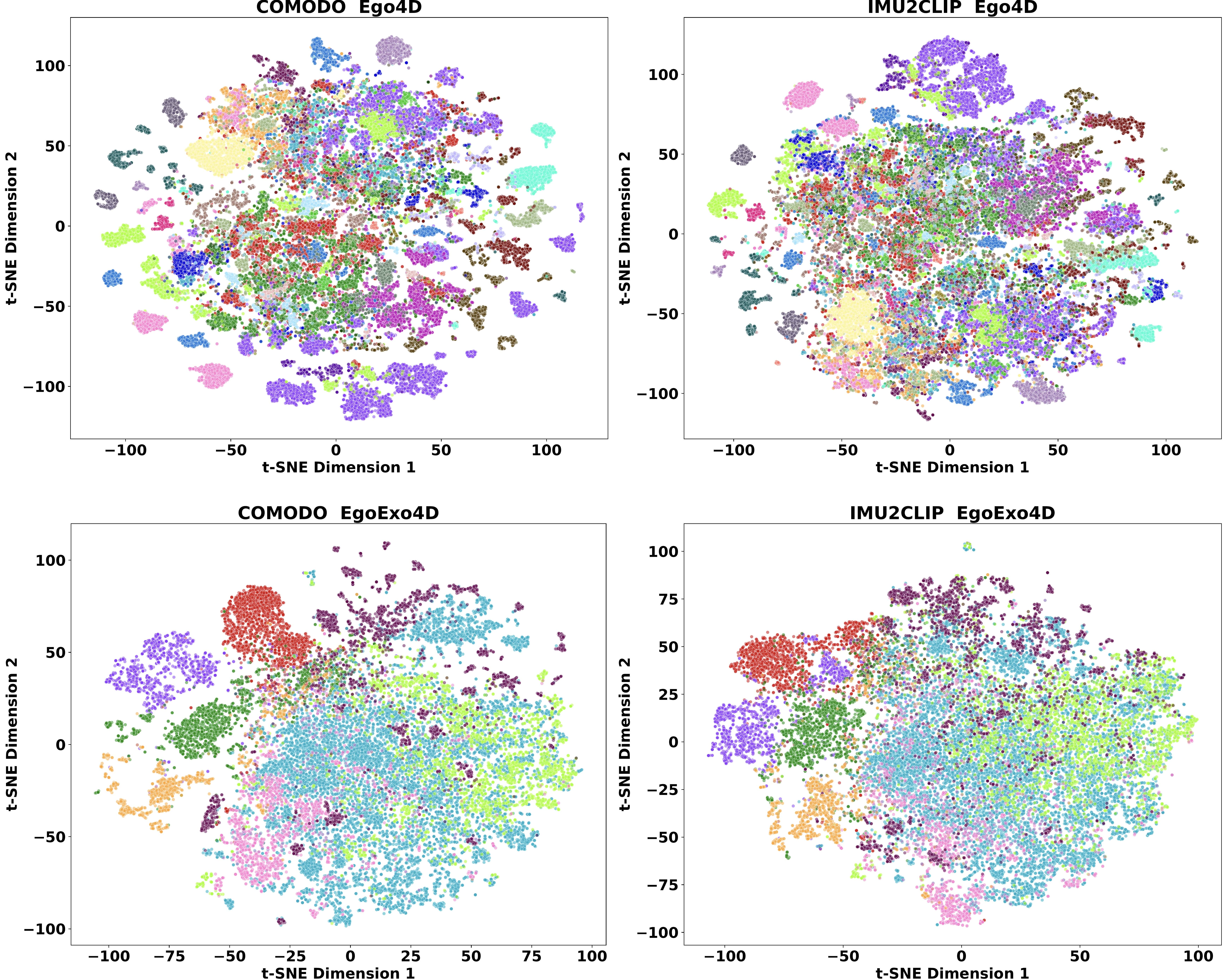}
   \caption{Qualitative t-SNE visualization.}
   \label{fig:tsne}
\end{figure}

\paragraph{Self-supervised methods.} We include two pretrained time-series foundation models, MOMENT-small~\cite{10.5555/3692070.3692712} and Mantis~\cite{feofanov2025mantis}. \revision{We also compare against CrossHAR~\cite{hong2024crosshar}, a hierarchical self-supervised pretraining framework for IMU-based HAR,} and IMU2CLIP \cite{moon2023imu2clip}, a contrastive cross-modal baseline.

\paragraph{Results.} As shown in Table~\ref{tab:results}, COMODO achieves state-of-the-art performance across all three datasets, ranking first on most metrics. On \textbf{Ego4D}, it outperforms IMU2CLIP by 4.3\% in A@1 and surpasses strong supervised models, Mantis and Attend and Discriminate, by 0.77\% and 1.01\%, respectively. \revision{COMODO also exceeds the performance of CrossHAR and IMUGPT 2.0 (Virtual+Real) by 8.17\% and 12.46\% in A@1}. On \textbf{EgoExo4D}, COMODO leads in A@1 and A@3, outperforming IMU2CLIP by 5.46\% and Mantis by 0.7\%, while achieving second-best in A@5, just narrowly behind Attend and Discriminate. \revision{In this dataset, COMODO consistently maintains a significant lead over CrossHAR and both variants of IMUGPT 2.0.} On \textbf{MMEA}, it exceeds IMU2CLIP by 1.07\% in A@1, ranks second to supervised Mantis by only 0.53\%, and achieves the best A@3 and A@5 scores. \revision{COMODO outperforms the finetuned CrossHAR by 7.51\% and demonstrates a substantial accuracy gain over IMUGPT 2.0.} Notably, COMODO consistently excels in A@3 and A@5 across all datasets, indicating its ability to capture richer semantic representations for a more accurate prediction distribution. These results validate the effectiveness of COMODO's cross-modal knowledge transfer, demonstrating its capacity to not only advance self-supervised learning but also match or surpass fully supervised models, underscoring its potential as a scalable, label-efficient solution for HAR. To qualitatively assess the learned representations, Figure~\ref{fig:tsne} shows t-SNE~\cite{maaten2008visualizing} visualizations on Ego4D and EgoExo4D. COMODO produces denser, more compact intra-class clusters with clearer inter-class boundaries, while IMU2CLIP exhibits greater overlap and dispersion.

\begin{table}[h!]
\centering
% \scriptsize
\begin{tabular}{cccccc}
\toprule
\textbf{Train} & \textbf{Test} & \textbf{Method} & \textbf{A@1} & \textbf{A@3} & \textbf{A@5} \\
\midrule
\revisionrow
\multirow{4}{*}{EgoExo4D} & \multirow{4}{*}{Ego4D} & CrossHAR & 30.55 & 55.95 & 69.70 \\
 &  & Mantis$^\dagger$ & 52.58 & 75.55 & 83.83 \\
 &  & IMU2CLIP & 51.75 & 74.44 & 83.05 \\
 &  & COMODO & \textbf{56.62} & \textbf{77.25} & \textbf{84.79} \\
\midrule
\revisionrow
\multirow{4}{*}{Ego4D} & \multirow{4}{*}{EgoExo4D} & CrossHAR & 68.22 & 95.11 & 98.64 \\
 &  & Mantis$^\dagger$ & 80.79 & 97.82 & 99.53 \\
 &  & IMU2CLIP & 79.70 & 97.68 & 99.45 \\
 &  & COMODO & \textbf{82.54} & \textbf{98.02} & \textbf{99.56} \\
\midrule
\revisionrow
\multirow{4}{*}{Ego4D} & \multirow{4}{*}{MMEA} & CrossHAR & 78.34 & 94.07 & 97.95 \\
 &  & Mantis$^\dagger$ & 91.72 & 98.48 & 99.32 \\
 &  & IMU2CLIP & 92.78 & 98.78 & 99.47 \\
 &  & COMODO & \textbf{94.22} & \textbf{98.78} & \textbf{99.62} \\
\midrule
\revisionrow
\multirow{4}{*}{EgoExo4D} & \multirow{4}{*}{MMEA} & CrossHAR & 79.86 & 94.45 & 97.72 \\
 &  & Mantis$^\dagger$ & 92.25 & 98.10 & 99.16 \\
 &  & IMU2CLIP & 92.78 & 98.93 & 99.70 \\
 &  & COMODO & \textbf{94.83} & \textbf{99.47} & \textbf{99.70} \\
\bottomrule
\end{tabular}
 \caption{\textbf{Cross-Dataset Generalization} results showing the method's ability to transfer to completely unseen datasets, where both the activity classes and devices differ from those in the training set. $\dagger$ indicates supervised fine-tuning on the source training dataset.}
\label{tab:transferability}
\end{table}

\subsection{Cross-Dataset Generalization}
The ultimate goal of representation learning is to learn features that are not only effective on the training domain but also generalize to new, unseen environments. Strong cross-dataset generalization is critical for deploying HAR models in diverse and realistic environments, where variations in sensor setups, activities, and user behaviors are common. To measure this capability, we designed a rigorous cross-dataset evaluation task. This task assesses how well the representations learned on one dataset (the source dataset) can be utilized on a completely different dataset (the target dataset) without additional fine-tuning, following the evaluation protocol in Section \ref{sec:inference} to assess out-of-domain generalization capability of the learned encoder. We experiment with the following setups: (1) EgoExo4D $\rightarrow$ Ego4D, (2) Ego4D $\rightarrow$ EgoExo4D, (3) Ego4D $\rightarrow$ MMEA, and (4) EgoExo4D $\rightarrow$ MMEA.

\paragraph{Results.} Table \ref{tab:transferability} presents the results of our cross-dataset generalization experiments. Notably, COMODO improves over IMU2CLIP, \revision{CrossHAR} and supervised fine-tuned Mantis baselines in all cases. In the EgoExo4D → Ego4D setup, COMODO outperforms IMU2CLIP by 4.87\% in A@1 while also surpassing the supervised fine-tuned Mantis by 4.04\%. \revision{Furthermore, COMODO demonstrates a substantial performance lead over CrossHAR (+26.07\% A@1), which suggests that intra-modal self-supervision struggles to bridge such large domain gaps without external semantic guidance.} When transferring from Ego4D to EgoExo4D, COMODO achieves the highest A@1 of 82.54\%, improving upon Mantis, IMU2CLIP, \revision{and CrossHAR} by 1.75\%, 2.84\%\revision{, and 14.32\%}. On the MMEA dataset, COMODO maintains its superiority, surpassing Mantis by 2.5\% and 2.58\% when trained on Ego4D and EgoExo4D, respectively. The findings from this cross-dataset evaluation are highly significant for practical HAR applications. They underscore COMODO is not merely improving performance on a specific data distribution and setting, but is fundamentally enhancing the representation quality and transferability of the IMU encoder, further demonstrating the effectiveness of cross-modal self-supervised knowledge transfer in enhancing IMU representation learning.

\subsection{\revision{Efficiency Analysis}}

\revision{
\paragraph{Inference Efficiency and Deployability} 
To validate the deployability of COMODO, we evaluate its inference efficiency in a single-threaded CPU setting on an AMD EPYC 9254 processor, simulating a resource-constrained environment. As illustrated in Figure~\ref{fig:efficiency}, COMODO achieves a superior balance between recognition accuracy and computational cost. All latency measurements are reported as the average per-sample inference time over 100 samples, and FLOPs are computed for a single forward pass.

\begin{figure}[h]
  \centering
   \includegraphics[width=\linewidth]{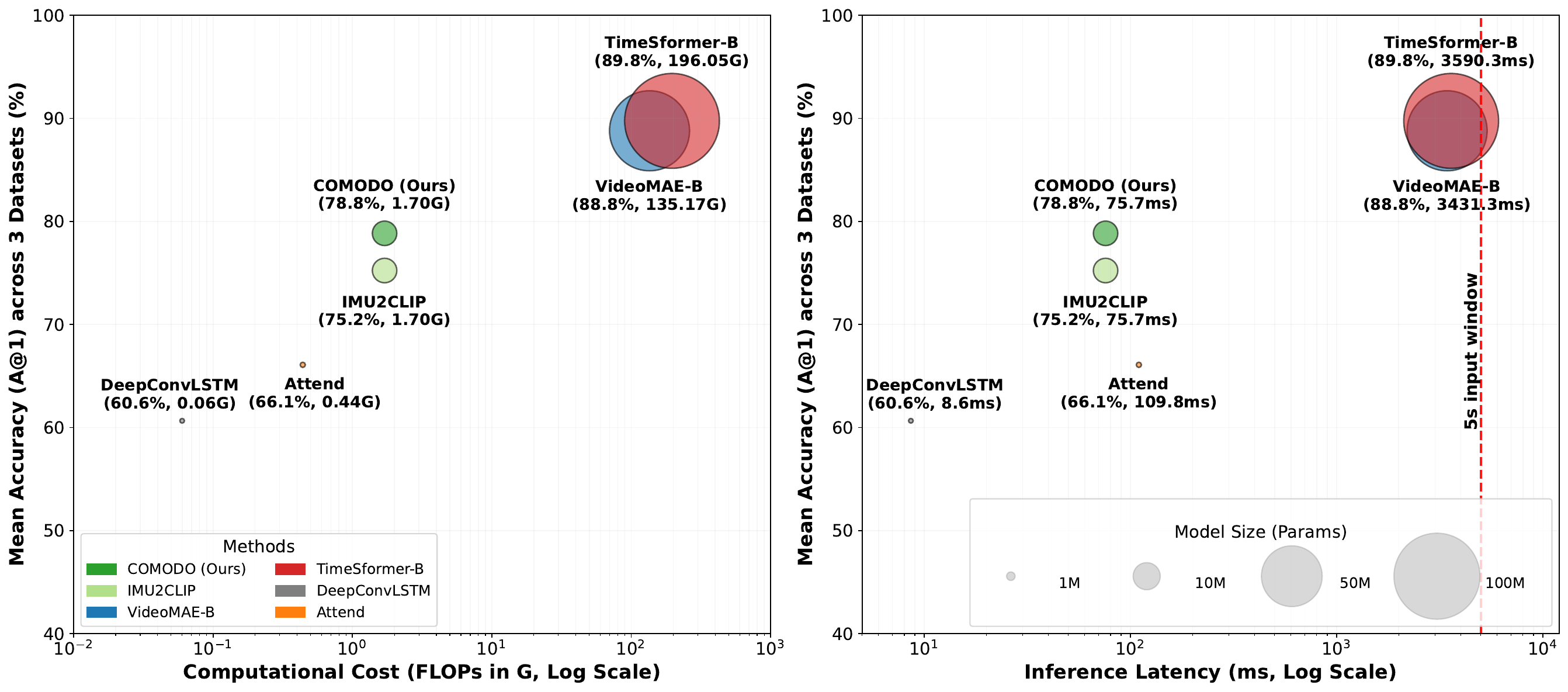}
   \caption{\revision{Comparison of performance and computational efficiency. We report mean A@1 accuracy across three datasets versus FLOPs (left) and inference latency (right), both in log scale. Bubble size denotes model size in terms of parameters. The red dashed line marks the 5-second input window.}}
   \label{fig:efficiency}
\end{figure}

Compared to high-performance video-based teachers (VideoMAE-Base-k400 and TimeSformer-Base-k400), COMODO offers a massive reduction in resource consumption. For instance, TimeSformer-Base-k400 (\#params: 121.4M) achieves 89.8\% accuracy but requires 196.05G FLOPs and 3590.3ms latency for a 5-second window. In contrast, COMODO (\#params: 8.1M) requires only 1.70G FLOPs and 75.7ms latency (47$\times$ speedup and 115$\times$ FLOPs reduction), while retaining a competitive accuracy of 78.8\%. As shown in Figure~\ref{fig:efficiency} (right), video models consume nearly 70\% of the window duration for inference, COMODO's 75.7ms latency represents only 1.5\% of the input duration.

Furthermore, compared to IMU-based models like DeepConvLSTM and Attend and Discriminate, COMODO delivers a substantial accuracy gain (up to +18.2\% mean A@1) while maintaining millisecond-level inference latency suitable for deployment. Moreover, despite sharing the same architecture and hardware footprint as IMU2CLIP, COMODO consistently achieves higher accuracy (+3.6\% mean A@1), demonstrating that similarity distribution distillation enhances semantic representations without introducing additional inference overhead.

\paragraph{Training Computational Cost and Scalability}
In addition to on-device inference efficiency, we also evaluate the training-time computational cost, which is critical for scalability on large datasets. Since COMODO relies on a frozen video teacher and only updates the lightweight IMU encoder, the training process is efficient. On a single NVIDIA L40S GPU, the average training time per epoch is 232.3s for Ego4D, 98.95s for EgoExo4D, and 11.8s for MMEA. Moreover, COMODO employs an offline feature extraction strategy (see Section~\ref{sec:training}), where video embeddings are pre-computed only once for each dataset and cached, and directly reused during IMU encoder training without invoking video inference. This design decouples the training-time computational cost from the complexity of the video teacher, enabling efficient training even when leveraging high-capacity video models. This low computational barrier ensures that COMODO is easily scalable to large datasets without requiring prohibitive computational resources.}

\subsection{Ablation Study}

\subsubsection{Different Teacher \& Student Networks.}
To evaluate the generalizability of COMODO across different models, we conduct experiments using two IMU student models (Mantis ~\cite{feofanov2025mantis} and MOMENT-small ~\cite{10.5555/3692070.3692712}) and two video teacher models (VideoMAE-Base-k400~\cite{tong2022videomae} and TimeSformer-Base-k400 ~\cite{bertasius2021space}). The results in Table~\ref{tab:diff_teacher_student} reveal that COMODO consistently and substantially outperforms the IMU2CLIP across every teacher-student configuration on all datasets.

\begin{table*}[!h]
 % \footnotesize
  \centering
  \setlength{\tabcolsep}{4.5pt}
    \begin{tabular}{@{}c c c | ccc | ccc | ccc@{}}
        \toprule
        \multirow{2}{*}{\textbf{IMU Model}} & \multirow{2}{*}{\textbf{Video Model}} & \multirow{2}{*}{\textbf{Methods}} & \multicolumn{3}{c|}{\textbf{Ego4D}} & \multicolumn{3}{c|}{\textbf{EgoExo4D}} & \multicolumn{3}{c}{\textbf{MMEA}} \\
        \cmidrule(lr){4-6} \cmidrule(lr){7-9} \cmidrule(lr){10-12}
         & & & \textbf{A@1} & \textbf{A@3} & \textbf{A@5} & \textbf{A@1} & \textbf{A@3} & \textbf{A@5} & \textbf{A@1} & \textbf{A@3} & \textbf{A@5} \\
        \midrule
        \multirow{2}{*}{MOMENT\(_S\)} & \multirow{2}{*}{VideoMAE\(_B\)} & IMU2CLIP & 41.39 & 65.82 & 76.45 & 69.77 & 94.61 & 98.81 & 78.42 & 91.19 & 94.60 \\
        \cmidrule(lr){3-12}
        & & COMODO & 57.74 & 76.66 & 83.79 & 81.60 & 97.68 & 99.48 & 79.48 & 91.19 & 94.45 \\
        \midrule
        \multirow{2}{*}{MOMENT\(_S\)} & \multirow{2}{*}{TimeSformer\(_B\)} & IMU2CLIP & 43.59 & 68.17 & 78.06 & 70.73 & 95.15 & 98.98 & 78.80 & 92.17 & 95.14 \\
        \cmidrule(lr){3-12}
        & & COMODO & 57.51 & 76.49 & 83.75 & 81.84 & 97.93 & 99.54 & 81.16 & 92.25 & 95.36 \\
        \midrule
        \multirow{2}{*}{Mantis} & \multirow{2}{*}{VideoMAE\(_B\)} & IMU2CLIP & 54.43 & 75.92 & 83.89 & 79.09 & 97.57 & 99.44 & 90.88 & 98.63 & 99.47 \\
        \cmidrule(lr){3-12}
        & & COMODO & \underline{59.00} & \textbf{78.98} & \textbf{85.95} & \underline{84.46} & \textbf{98.31} & \underline{99.56} & \textbf{93.54} & \textbf{99.24} & \textbf{99.77} \\
        \midrule
        \multirow{2}{*}{Mantis} & \multirow{2}{*}{TimeSformer\(_B\)} & IMU2CLIP & 54.83 & 76.04 & 84.03 & 79.46 & 97.68 & 99.44 & 91.41 & 98.71 & 99.47 \\
        \cmidrule(lr){3-12}
        & & COMODO & \textbf{59.13} & \underline{78.79} & \underline{85.65} & \textbf{84.92} & \underline{98.28} & \textbf{99.59} & \underline{92.48} & \underline{99.01} & \textbf{99.77} \\
        \bottomrule
    \end{tabular}
  \caption{Result for different video teacher-IMU student model combinations on Ego4D, EgoExo4D, and MMEA. The best are indicated in \textbf{bold}, and the second-best are \underline{underlined}.}
  \label{tab:diff_teacher_student}
\end{table*}

We observe two key trends. First, regarding the student models, Mantis consistently proves to be a more capable learner than MOMENT-small, achieving significantly higher performance regardless of the teacher. This aligns with its superior standalone results in Table~\ref{tab:results}. Second, regarding the video teacher, we observe that both VideoMAE-Base-k400 and TimeSformer-Base-k400 act as highly effective knowledge sources. While neither demonstrates universal superiority across all settings, their standalone performances are strong and highly comparable. This highlights a crucial strength of our method: COMODO's effectiveness is not contingent on a specific teacher architecture, demonstrating strong model-agnostic capabilities and adaptability to both video and time-series models. This is further evidenced by the fact that the combination of the stronger Mantis student with either teacher consistently yields the best and second-best results overall.

\paragraph{Video Teacher Performance} To contextualize the performance of our IMU-only model, we present the capabilities of the video teacher models in Table~\ref{tab:video_teacher}. This serves as a proxy for the intrinsic difficulty of each classification task. We report two sets of teacher results: pre-trained models (VideoMAE\(_B\)-k400 and TimeSformer\(_B\)-k400) evaluated by training an SVM classifier on their extracted features, which aligns with the evaluation setting for our student model in the main paper. Additionally, we include results for these same models after being supervised fine-tuned on each training dataset \(\dag\), evaluated using their classifier heads setting.

\begin{table*}[h]
  \centering
    \begin{tabular}{@{}p{3.5cm} | ccc | ccc | ccc@{}}
        \toprule
        \multirow{2}{*}{\textbf{Methods}} & \multicolumn{3}{c|}{\textbf{Ego4D}} & \multicolumn{3}{c|}{\textbf{EgoExo4D}} & \multicolumn{3}{c}{\textbf{MMEA}} \\
        \cmidrule(lr){2-4} \cmidrule(lr){5-7} \cmidrule(lr){8-10}
        & \textbf{A@1} & \textbf{A@3} & \textbf{A@5} & \textbf{A@1} & \textbf{A@3} & \textbf{A@5} & \textbf{A@1} & \textbf{A@3} & \textbf{A@5} \\
        \midrule
        \raggedright COMODO (\textbf{Ours}) & 59.13 & 78.79 & 85.65 & 84.92 & 98.28 & 99.59 & 92.48 & 99.01 & 99.77 \\
        \midrule
        \midrule
        \raggedright VideoMAE\(_B\)-k400 & 74.31 & 87.80 & 92.15 & 99.27 & 99.90 & 100 & 92.78 & 99.32 & 99.62 \\
        \midrule
        \raggedright TimeSformer\(_B\)-k400 & 76.16 & 88.99 & 92.43 & 99.63 & 99.96 & 100 & 93.47 & 99.01 & 99.54 \\
        \midrule
        \raggedright VideoMAE\(_B\)-k400$\dag$ & 77.99 & 86.75 & 89.53 & 99.81 & 99.98 & 99.99 & 93.50 & 98.66 & 99.47\\
        \midrule
        \raggedright TimeSformer\(_B\)-k400$\dag$ & 79.56 & 88.18 & 90.47 & 99.98 & 100 & 100 & 93.73 & 98.43 & 99.38 \\
        \bottomrule
    \end{tabular}
  \caption{Performance comparison between the IMU student (COMODO) and its video teachers. $\dag$ denotes supervised finetuned on the training dataset.}
  \label{tab:video_teacher}
\end{table*}

The analysis reveals that the video teachers achieve near-perfect accuracy on EgoExo4D and MMEA, signifying that the activities in these datasets are highly distinguishable from the visual modality. In contrast, the lower scores on Ego4D confirm its greater complexity. Impressively, our IMU-only COMODO model successfully closes a significant portion of the modality gap, especially on the more challenging Ego4D dataset. On EgoExo4D and MMEA, COMODO's performance is remarkably close to that of its video teachers, demonstrating that our distillation framework effectively transfers rich semantic knowledge from the powerful video modality to the far more lightweight IMU modality in a self-supervised way.

\subsubsection{FIFO Queue Size}
The optimal queue size \(|Q|\) is dataset-specific, as different datasets benefit from different queue configurations. Our ablation study (Figure \ref{fig:queue_size}) reveals that increasing queue size does not always yield better results. Instead, the relationship between queue size and performance is influenced by the characteristics of the dataset:

\begin{figure}[!h]
  \centering
   \includegraphics[width=1\linewidth]{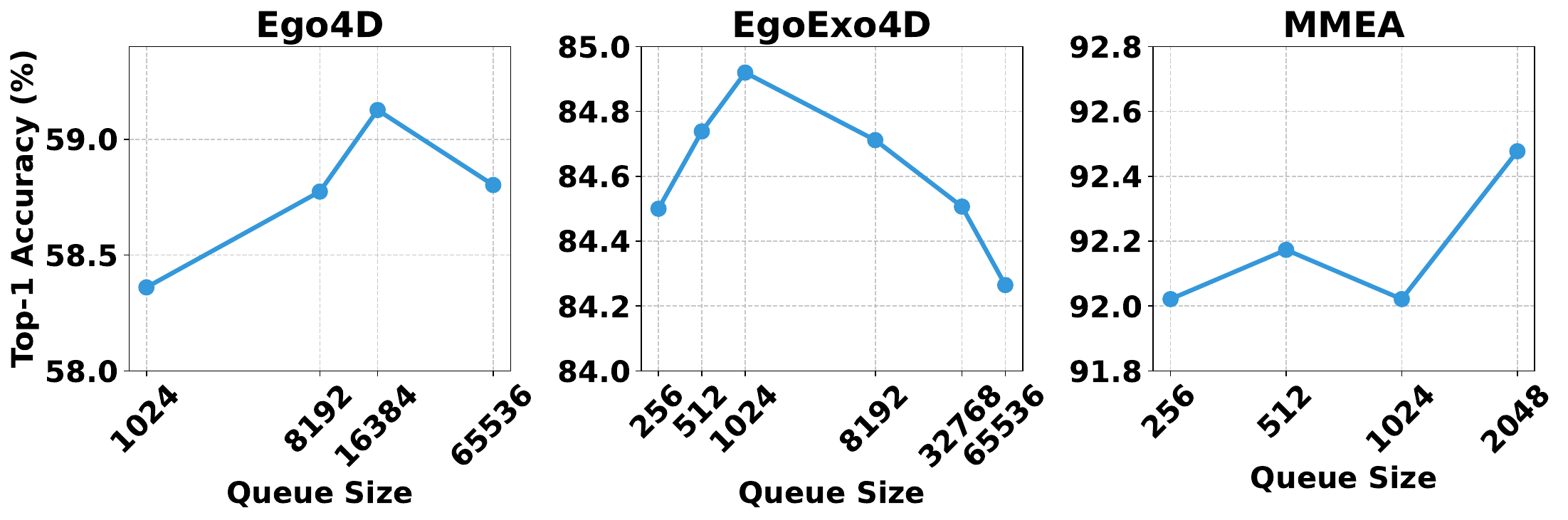}
   \caption{Impact of queue size on accuracy across datasets.}
   \label{fig:queue_size}
\end{figure}

Ego4D (\(|Q|\) = 16384): As a large-scale dataset with diverse activities (31 classes), increasing the queue size generally improves performance by providing a richer set of negative samples for contrastive learning. However, we observed that the largest queue size tested (65536) does not yield the best performance. Prior research ~\cite{limkonchotiwat2022congen} also suggests that the optimal queue size is not necessarily the largest, and an overly large queue may not always lead to improved results.

EgoExo4D (\(|Q|\) = 1024): Empirical results indicate that increasing the queue size initially improves performance, but beyond \(|Q|\) = 1024, further increasing the queue size leads to a decline in accuracy (Figure \ref{fig:queue_size}). Unlike Ego4D and MMEA, EgoExo4D has only 8 activity classes, leading to highly concentrated feature distributions. When the queue size is large, excessive redundant negatives from the same limited set of categories create sharp similarity spikes in the distribution. This makes it harder for the student model to align with the teacher’s representation. Prior work ~\cite{gu2024minillm} suggests that KL divergence is easier to minimize when the distribution has a single dominant mode, whereas in the distributions of multiple distinct modes, forward KL forces the model to cover all modes, potentially assigning probability mass to less relevant regions. A moderate queue size helps regulate the number of distinct similarity peaks, stabilizing training and improving teacher-student alignment.

MMEA (\(|Q|\) = 2048): This dataset contains 32 activity classes but has significantly fewer IMU samples compared to the other datasets. Given the limited number of samples, a very large queue is impractical. Instead, selecting a queue size that is proportionally appropriate for the dataset ensures that each mini-batch includes sufficiently diverse negatives. Our experiments show that a queue size of 2048 achieves the best performance on MMEA (Figure \ref{fig:queue_size}). Based on this observation, we choose 2048 as the queue size, which is relatively large for this dataset but helps maintain a balance between negative sample diversity and dataset constraints.

\subsubsection{Pooling Methods}
\label{sec:pooling-methods}
We compare mean pooling and concat pooling to evaluate their impact on IMU representation learning. Mean pooling compresses features by averaging across all sensor channels, while concat pooling preserves channel-wise distinctions by concatenating the features from each channel. For this ablation study, we use MOMENT-small as the IMU student model, as it provides built-in support for both pooling methods. In contrast, Mantis is designed with concat pooling as its default architecture, making it less straightforward to modify for this comparison.

\begin{table}[h]
\centering
\begin{tabular}{ccccc}
\toprule
\textbf{Train} & \textbf{Method} & \textbf{A@1} & \textbf{A@3} & \textbf{A@5} \\
\midrule
\multirow{2}{*}{Ego4D} & Mean & 55.69 & 75.21 & 82.92 \\
 & Concat & \textbf{57.51} & \textbf{76.49} & \textbf{83.75} \\
\midrule
\multirow{2}{*}{EgoExo4D} & Mean & 80.86 & 97.48 & 99.42 \\
 & Concat & \textbf{81.84} & \textbf{97.93} & \textbf{99.54} \\
\midrule
\multirow{2}{*}{MMEA} & Mean & 74.32 & 88.91 & 93.09 \\
 & Concat & \textbf{81.16} & \textbf{92.25} & \textbf{95.36} \\
\bottomrule
\end{tabular}
\caption{Comparison of mean pooling and concat pooling for IMU representation learning across different datasets. }
\label{tab:pooling}
\end{table}

Table \ref{tab:pooling} shows that concat pooling consistently outperforms mean pooling across all datasets. On Ego4D, it improves A@1 by 1.82\%, while on EgoExo4D, the gain is 0.98\%. The most significant improvement is observed on MMEA, where A@1 increases by 6.84\%, indicating that maintaining per-channel information is especially beneficial for this dataset. These results suggest that mean pooling overly compresses features, leading to information loss, while concat pooling retains richer representations, enhancing classification accuracy. Given its superior performance across all datasets, we adopt concat pooling as the default method in our framework, as discussed in Section \ref{sec:training}.

\subsubsection{Distillation Methods.}
\label{sec:ablation_distillation}
To \revision{empirically validate the theoretical analysis presented in Section \ref{sec:theoretical_diff}}, we compare our strategy against two representative baseline methods.

(1) L2 Distance Minimization ~\cite{DBLP:journals/corr/RomeroBKCGB14}: L2 distance minimization directly aligns the student and teacher representations by minimizing the squared L2 distance between the IMU student embedding \(z_{xi}\) and the video teacher embedding \(z_{vi}\) for an identical paired video-IMU sample (\(v_i, x_i\)). This method assumes that the feature spaces of different modalities are inherently aligned. As a result, L2 minimization can lead to suboptimal feature alignment.

(2) InfoNCE-based Contrastive Learning: InfoNCE optimizes the student encoder by maximizing the similarity between positive video-IMU pairs while minimizing similarity with randomly sampled negative instances. \revision{As discussed in Section~\ref{sec:theoretical_diff}, this formulation corresponds to the ``hard'' contrastive objective, which imposes a one-hot target distribution.} While effective in learning discriminative features, this method treats all negative samples equally, potentially misclassifying semantically similar negatives, which limits its adaptability in cross-modal distillation.

\begin{figure}[!h]
  \centering
   \includegraphics[width=1\linewidth]{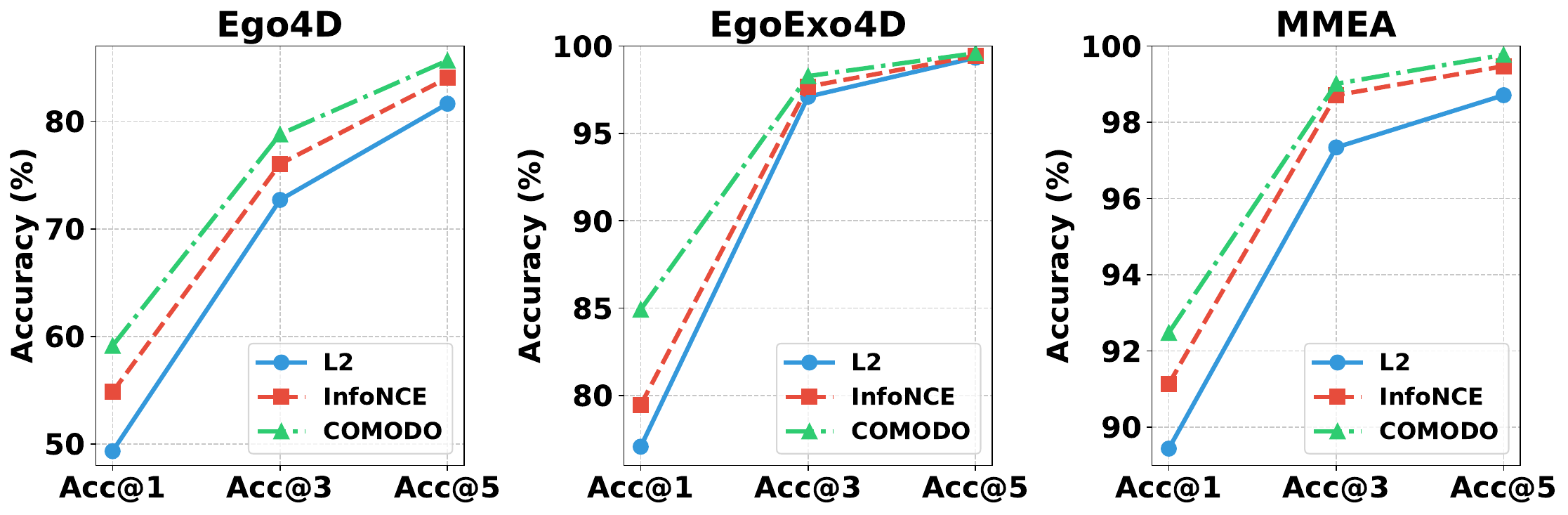}
   \caption{Accuracy of distillation methods across datasets.}
   \label{fig:distillation_comparison}
\end{figure}

(3) COMODO (ours): COMODO extends contrastive learning by distilling the \emph{similarity distribution} rather than relying solely on instance-wise alignment. \revision{Consistent with our gradient analysis in Eq.~\ref{eq:gradient_analysis}}, unlike InfoNCE, which assumes all negatives are equally dissimilar, COMODO preserves the structure of the teacher’s similarity space, leading to more robust and generalizable feature representations. Figure \ref{fig:distillation_comparison} shows that COMODO consistently outperforms L2 and InfoNCE across all datasets, demonstrating the effectiveness of similarity distribution distillation in cross-modal representation learning.

% (1) L2 Distance Minimization\cite{DBLP:journals/corr/RomeroBKCGB14} directly aligns the student and teacher by minimizing the squared L2 distance between the IMU student embedding \(z_{x i}\) and the video teacher embedding \(z_{v i}\) for paired samples \((v_i, x_i)\). This assumes inherently compatible feature spaces, which can yield suboptimal alignment.

% (2) InfoNCE-based Contrastive Learning optimizes the student by increasing similarity for positive video–IMU pairs and decreasing it for negatives samples. Although effective for discrimination, it treats all negatives uniformly and can penalize semantically similar negatives, limiting adaptability in cross-modal distillation.

% (3) COMODO (ours) extends contrastive learning by distilling the \emph{similarity distribution} rather than relying on instance-level matches. By preserving the teacher's similarity structure instead of assuming all negatives are equally dissimilar, COMODO produces more robust and generalizable representations. As shown in Figure~\ref{fig:distillation_comparison}, COMODO consistently outperforms L2 and InfoNCE across all datasets.

\subsubsection{Sampling Methods.}
\label{sec:sampling_method}
To investigate the effectiveness of the FIFO queue in our cross-modal distillation framework, we conduct an ablation study comparing it against random sampling strategy. For a fair comparison, all other settings, including the number of samples used for distillation in each step (i.e., the queue size for FIFO and the number of samples for random selection), are kept identical.

% To investigate the FIFO queue in our cross-modal distillation, we ablate it against random sampling. For fairness, all other settings are identical, including the number of samples per step (i.e., the queue size for FIFO and the number of samples for random selection), with a detailed ablation on queue size provided in the Appendix.

(1) Empirical Results. As shown in Table \ref{tab:sampling}, FIFO yields superior performance across all datasets. This consistent improvement motivates a deeper investigation into the optimization dynamics induced by the sampling method.

\begin{table}[htbp]
\centering
\begin{tabular}{ccccc}
\toprule
\textbf{Train} & \textbf{Sampling} & \textbf{A@1} & \textbf{A@3} & \textbf{A@5} \\
\midrule
\multirow{2}{*}{Ego4D} & FIFO & \textbf{59.13} & \textbf{78.79} & \textbf{85.65} \\
 & Random & 58.47 & 78.72 & 85.64 \\
\midrule
\multirow{2}{*}{EgoExo4D} & FIFO & \textbf{84.92} & 98.28 & \textbf{99.59} \\
 & Random & 84.68 & \textbf{98.32} & 99.59 \\
\midrule
\multirow{2}{*}{MMEA} & FIFO & \textbf{92.48} & \textbf{99.01} & \textbf{99.77} \\
 & Random & 92.25 & 98.63 & 99.70 \\
\bottomrule
\end{tabular}
\caption{Comparison of FIFO and random sampling.}
\label{tab:sampling}
\end{table}

(2) Theoretical Hypothesis. Sampling as an implicit temporal prior on the minimization flow. We posit that the choice of sampling strategy functions as an implicit regularizer, imposing a distinct prior on the optimization process's minimization flow. The random sampling approach draws a mini-batch of video embeddings randomly and independently from the entire dataset at each iteration. This results in gradient estimations that are largely independent from one step to the next, potentially leading to a high-variance, discontinuous optimization path. In contrast, the FIFO queue introduces a temporal continuity prior. At any two consecutive steps, \(t\) and \(t+1\), the queue contents \(Q_t\) and \(Q_{t+1}\) are highly correlated, as only the oldest batch of embeddings is replaced by the newest. This ensures that the data distribution used to compute the distillation loss, \(P_v(i|Q)\), transitions smoothly over time. Consequently, the loss landscape ``seen'' by the student model evolves gradually, encouraging a more stable and continuous minimization flow.

% (3) Validation and Discussion. Our investigation is conceptually motivated by ~\citet{charpiat2007generalized}, who showed that modifying the inner product space for gradient computation can introduce priors that yield smoother optimization paths. While their work focuses on a gradient-level prior by altering the gradient, we explore an alternative: shaping the optimization dynamics by structuring the data distribution, which we term a data-level prior. To validate that this prior indeed leads to a more continuous optimization flow, we design an experiment to visualize the path's continuity. We compute the gradient of \(L_{CE}\) with respect to the IMU student model's parameters \(\theta_x\) at each iteration \(t\), denoted as \(\mathbf{g}_t = \nabla_{\theta_x} L_{CE}^{(t)}\). We then measure the cosine similarity between the current gradient \(\mathbf{g}_t\) and the moving average of the \(K\) preceding gradients, \(\bar{\mathbf{g}}_{t-1} = \frac{1}{K} \sum_{k=1}^{K} \mathbf{g}_{t-k}\). The results are visualized in Figure~\ref{fig:gradient}, with an ablation on different \(K\) values included in the Appendix.

\begin{figure}[h!]
  \centering
  %----- K=30 Row -----
  \begin{subfigure}[b]{0.48\linewidth}
    \centering
    \includegraphics[width=\linewidth]{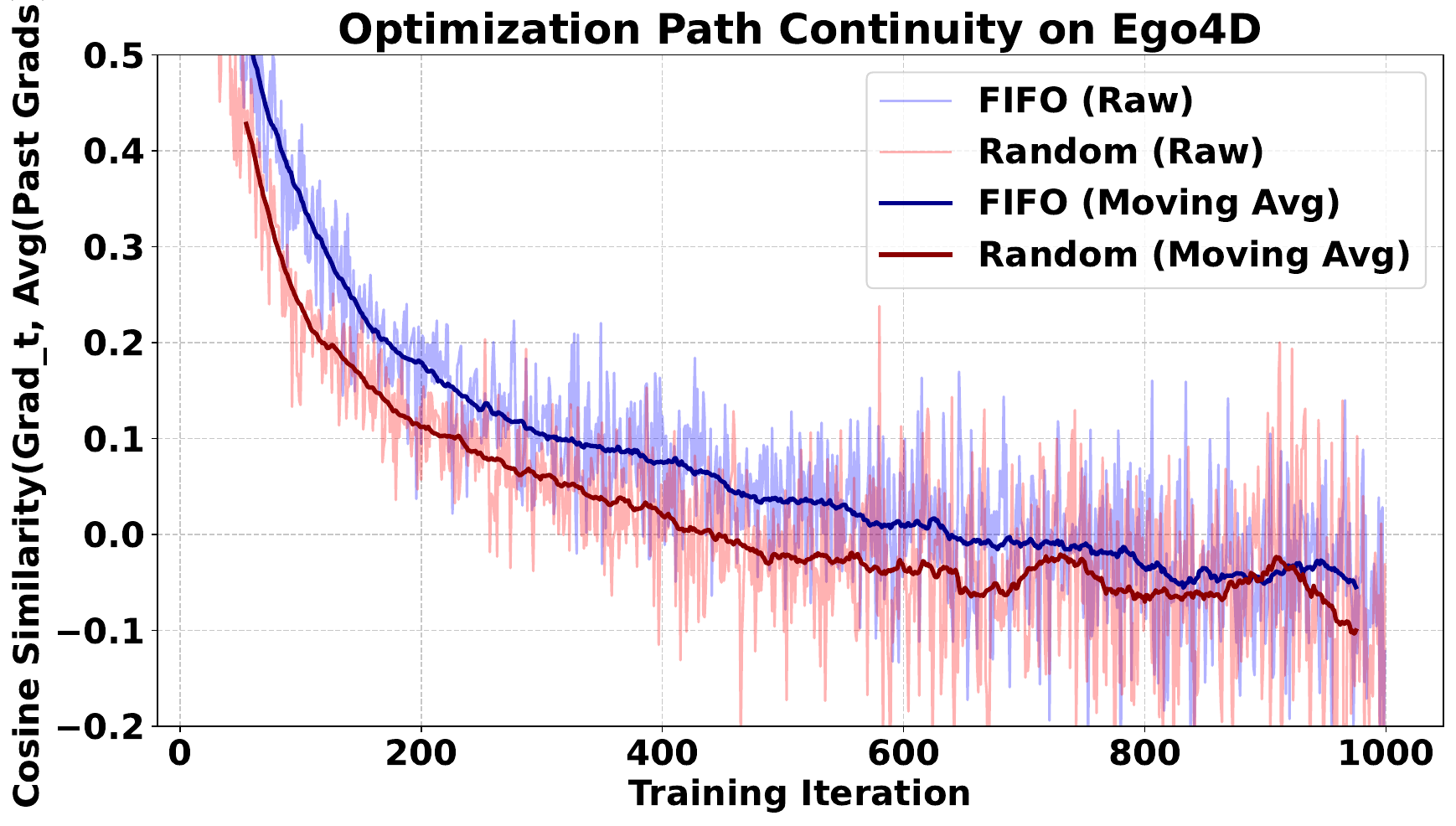}
    \caption{Ego4D, K=30}
    \label{fig:grad_ego4d_30}
  \end{subfigure}
  \hfill
  \begin{subfigure}[b]{0.48\linewidth}
    \centering
    \includegraphics[width=\linewidth]{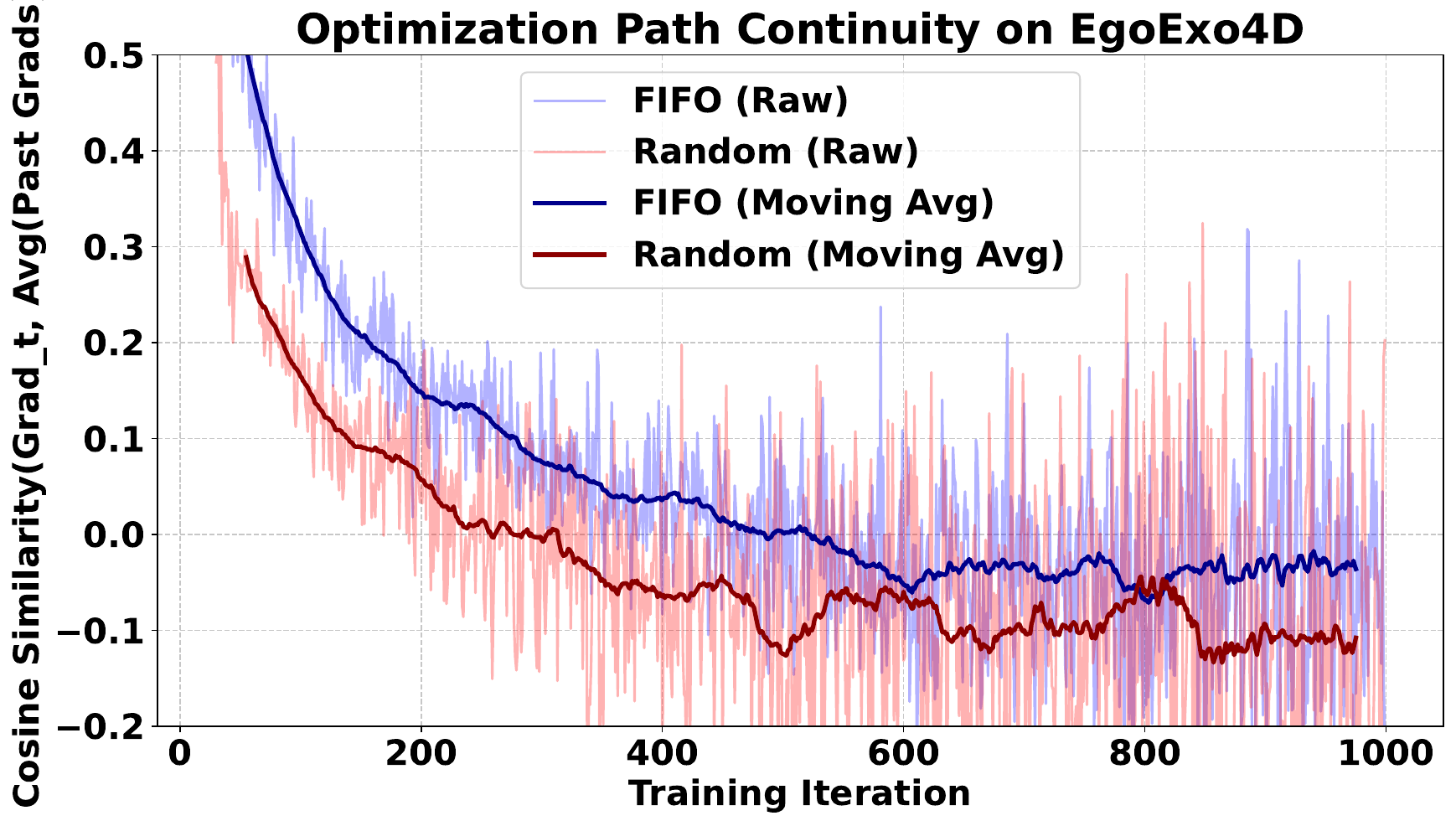}
    \caption{EgoExo4D, K=30}
    \label{fig:grad_egoexo_30}
  \end{subfigure}

  \vspace{0.3cm} % Add some vertical space between rows
  
  %----- K=50 Row -----
  \begin{subfigure}[b]{0.48\linewidth}
    \centering
    \includegraphics[width=\linewidth]{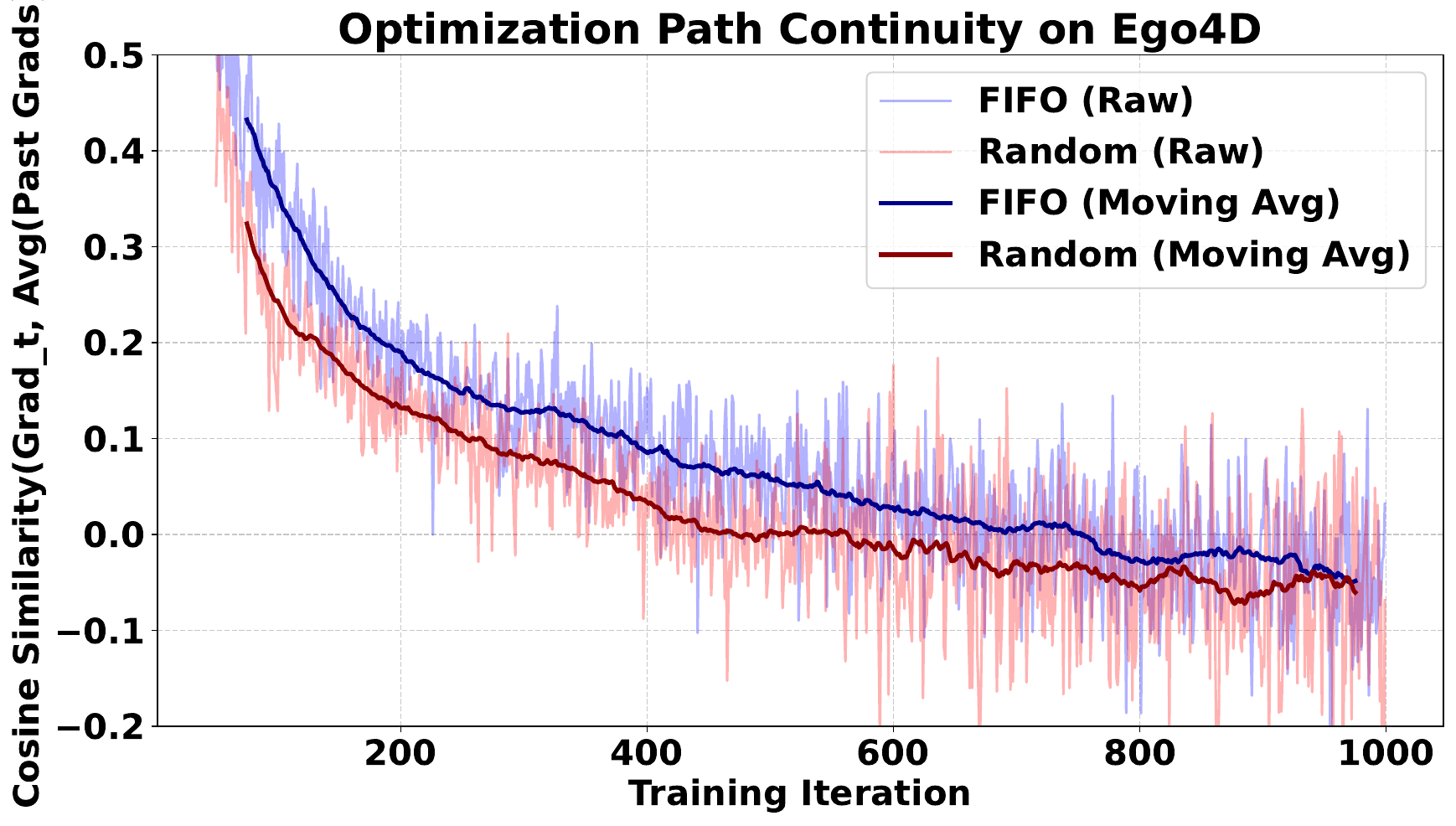}
    \caption{Ego4D, K=50}
    \label{fig:grad_ego4d_50}
  \end{subfigure}
  \hfill
  \begin{subfigure}[b]{0.48\linewidth}
    \centering
    \includegraphics[width=\linewidth]{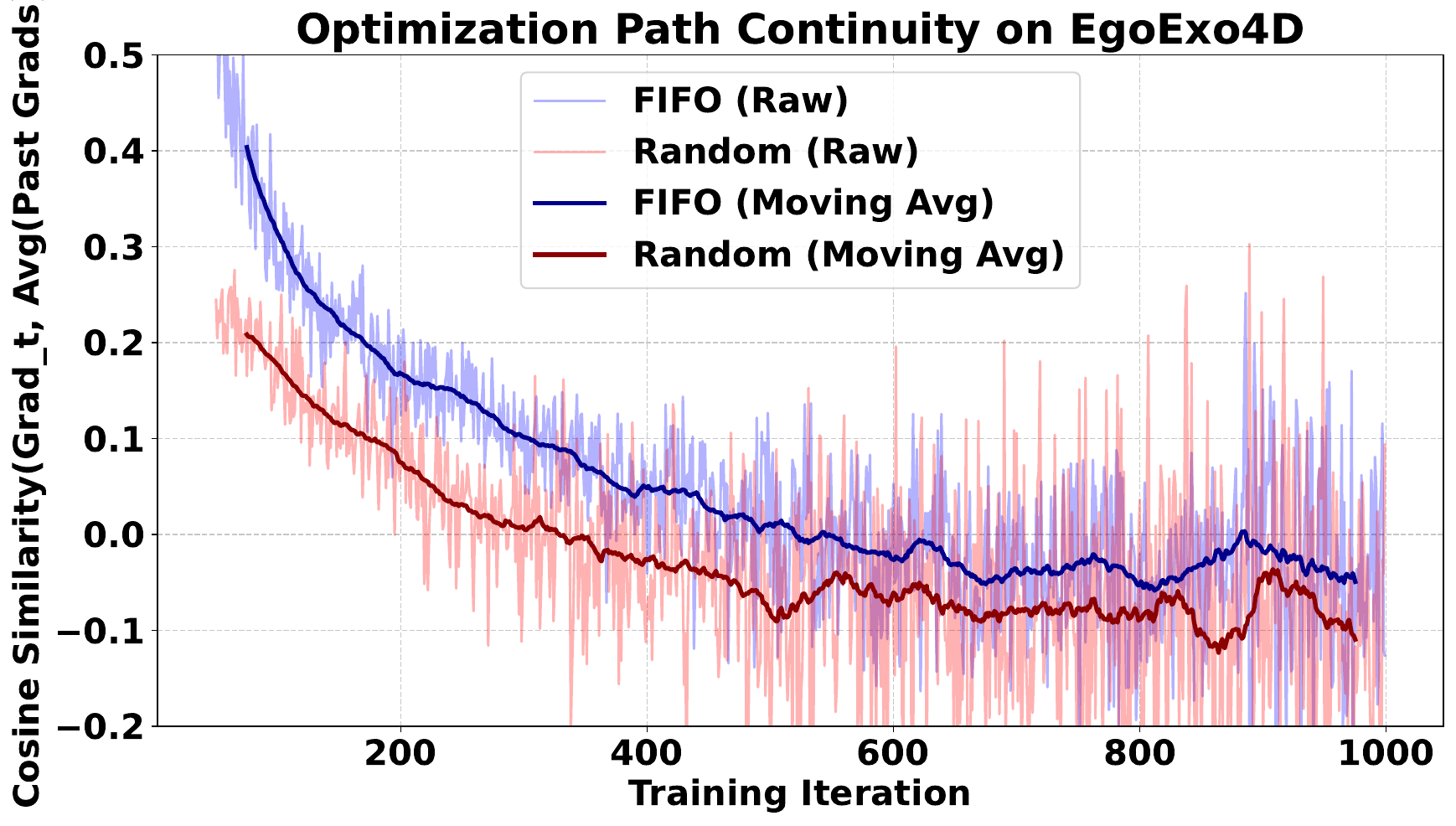}
    \caption{EgoExo4D, K=50}
    \label{fig:grad_egoexo_50}
  \end{subfigure}
  
  \vspace{0.3cm} % Add some vertical space between rows

  %----- K=100 Row -----
  \begin{subfigure}[b]{0.48\linewidth}
    \centering
    \includegraphics[width=\linewidth]{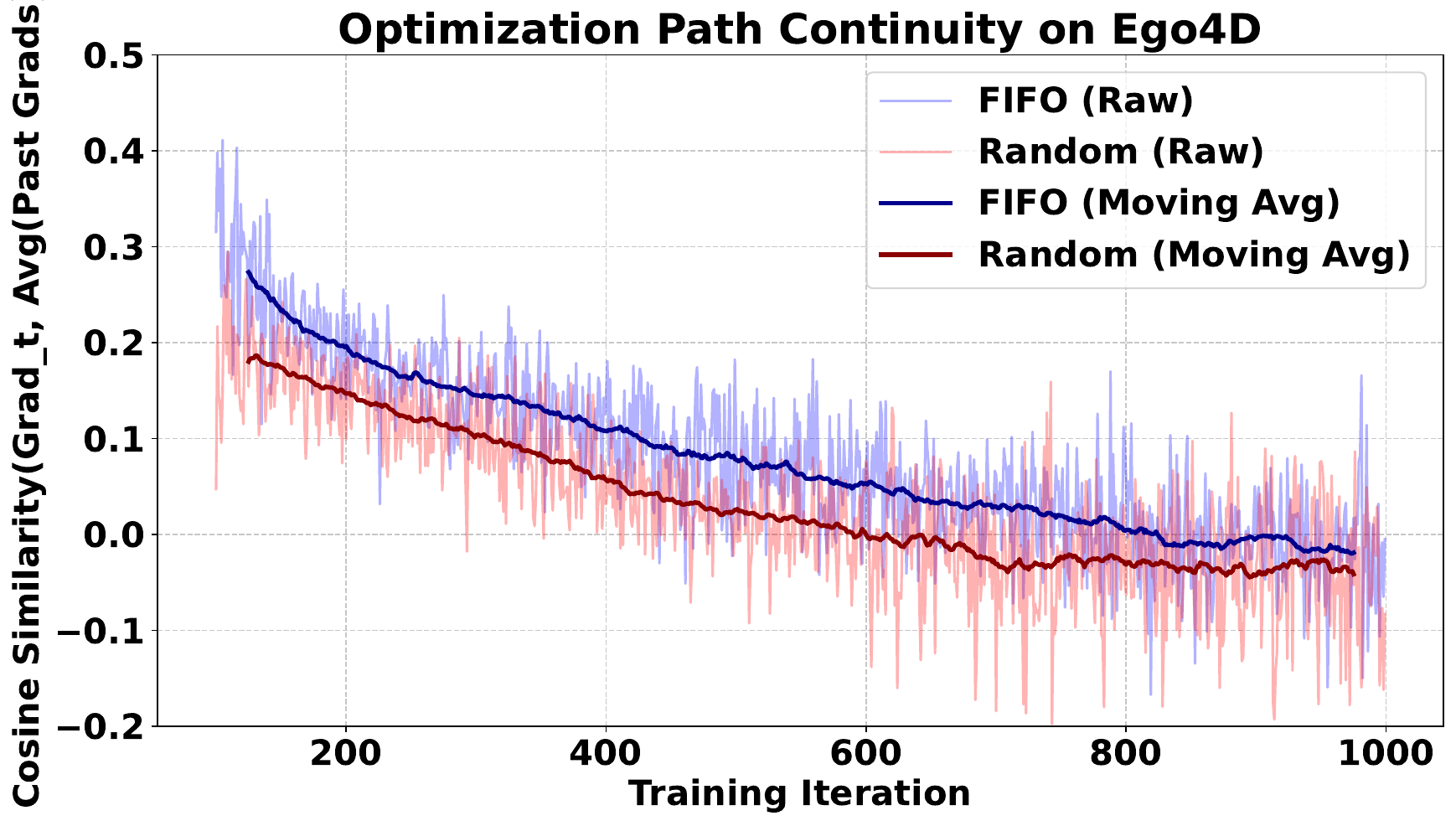}
    \caption{Ego4D, K=100}
    \label{fig:grad_ego4d_100}
  \end{subfigure}
  \hfill
  \begin{subfigure}[b]{0.48\linewidth}
    \centering
    \includegraphics[width=\linewidth]{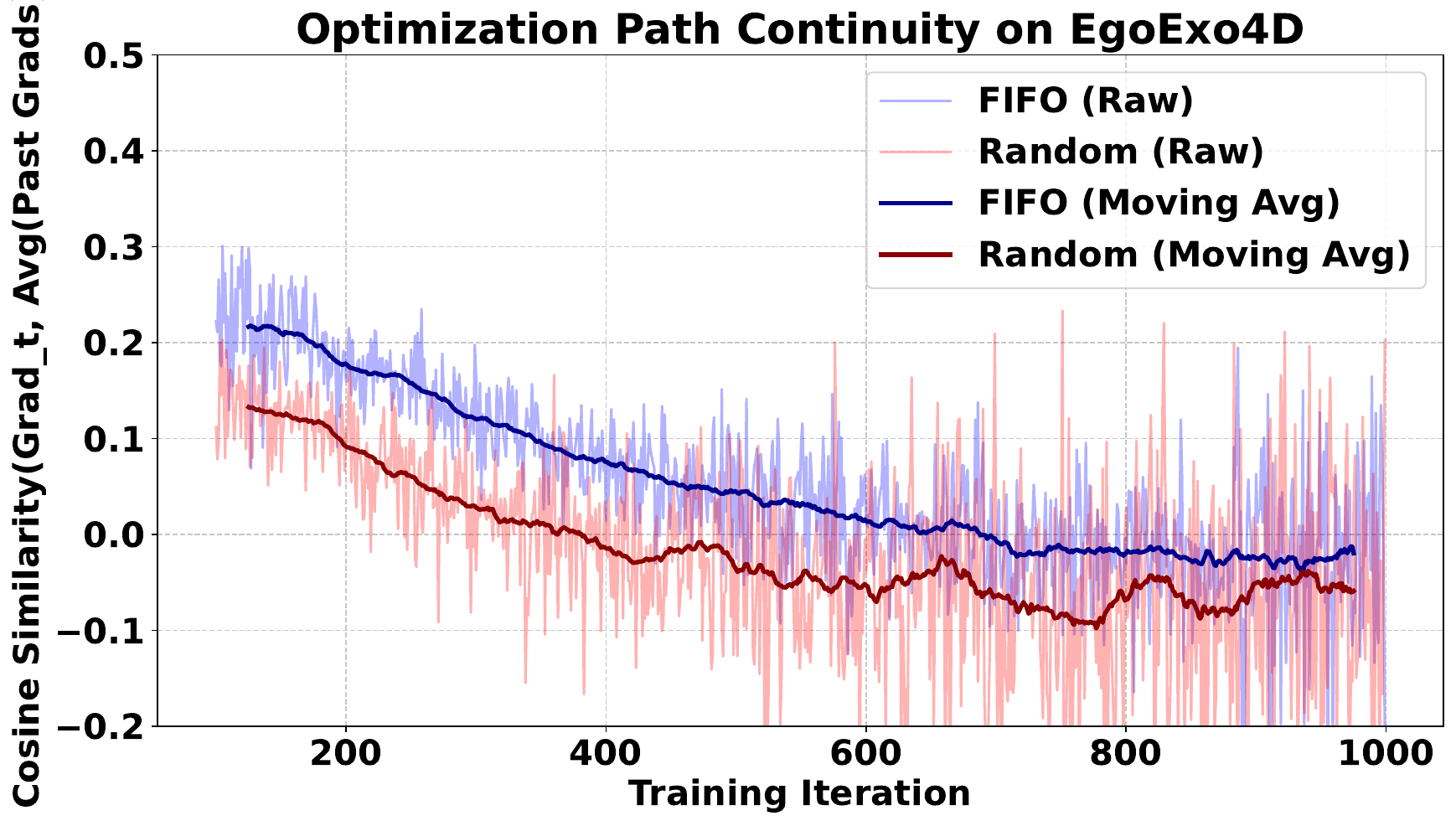}
    \caption{EgoExo4D, K=100}
    \label{fig:grad_egoexo_100}
  \end{subfigure}

  \caption{Comparison on the different moving average window size \(K\) for the gradient path continuity for FIFO (blue) and Random (red) sampling.}
  \label{fig:gradient}
\end{figure}

(3) Validation and Discussion. Our investigation is conceptually motivated by ~\citet{charpiat2007generalized}, who showed that modifying the inner product space for gradient computation can introduce priors that yield smoother optimization paths. While their work focuses on a gradient-level prior by altering the gradient, we explore an alternative: shaping the optimization dynamics by structuring the data distribution, which we term a data-level prior. To validate that this prior indeed leads to a more continuous optimization flow, we design an experiment to visualize the path's continuity. We compute the gradient of \(L_{CE}\) with respect to the IMU student model's parameters \(\theta_x\) at each iteration \(t\), denoted as \(\mathbf{g}_t = \nabla_{\theta_x} L_{CE}^{(t)}\). We then measure the cosine similarity between the current gradient \(\mathbf{g}_t\) and the moving average of the preceding \(K\) gradients, \(\bar{\mathbf{g}}_{t-1} = \frac{1}{K} \sum_{k=1}^{K} \mathbf{g}_{t-k}\). This metric quantifies the consistency of the gradient direction over time; a higher cosine similarity indicates a more stable and coherent optimization trajectory.

To ensure the robustness of our findings, we conducted this analysis across different window sizes for the moving average, specifically for \(K \in \{30, 50, 100\}\). A smaller \(K\) makes the measure more sensitive to high-frequency changes in gradient direction, while a larger \(K\) provides a more smoothed, long-term estimate of the path's direction.

% \begin{figure}[ht]
%   \centering
%   \begin{subfigure}[b]{0.48\linewidth}
%     \centering
%     \includegraphics[width=\linewidth]{sec/images/gradient_continuity_Ego4D_50.pdf}
%     \label{fig:grad_ego4d}
%   \end{subfigure}
%   \hfill
%   \begin{subfigure}[b]{0.48\linewidth}
%     \centering
%     \includegraphics[width=\linewidth]{sec/images/gradient_continuity_EgoExo4D_50.pdf}
%     \label{fig:grad_egoexo}
%   \end{subfigure}
%   \caption{Comparison of gradient path continuity under FIFO and Random sampling strategies.}
%   \label{fig:gradient}
% \end{figure}

The results, visualized in Figure~\ref{fig:gradient}, provide compelling and consistent evidence supporting our hypothesis. Across both the Ego4D (left column) and EgoExo4D (right column) datasets, the gradient continuity for the FIFO strategy (blue curves) is markedly consistently higher than that of random sampling (red curves). This superiority of FIFO holds true for all tested window sizes of \(K\), while the absolute smoothness of the moving average curves naturally increases with a larger \(K\) (as seen by the less jagged lines in the bottom row \(K=100\) compared to the top \(K=30\)).

This robustness analysis confirms that the temporal continuity prior induced by the FIFO queue is not a mere artifact of a specific hyperparameter choice but a fundamental mechanism stabilizing the optimization process. The sustained higher cosine similarity indicates that the FIFO queue induces a more continuous and stable minimization flow, where the optimization trajectory is more coherent over time. We posit that this stabilized gradient direction allows the optimizer to converge more efficiently and effectively, preventing unstable updates and ultimately explaining the superior performance of the FIFO strategy observed in our experiments.
\section{Discussion and Implications}
Our work demonstrates that cross-modal, self-supervised distillation from video to IMU is a highly effective and scalable paradigm for overcoming the long-standing data scarcity problem in wearable HAR. The success of COMODO not only presents a state-of-the-art technical solution but also offers several broader implications for the design and development of future human-centered ubiquitous technologies.

\subsection{Implications for Egocentric HAR: A Path Beyond Manual Annotation}
A key implication of COMODO is the shift it suggests in the data collection and development lifecycle for wearable HAR systems.  The field has long been hampered by the annotation bottleneck, a challenge frequently cited as a primary obstacle to building robust, generalizable real-world ("in-the-wild") models~\cite{haresamudram2025past}. COMODO offers a practical solution to this long-standing dilemma.

First, our framework acts as a conduit to leverage the immense progress and rich semantic knowledge encapsulated in large-scale, pre-trained vision models. The computer vision community has invested enormous effort in creating powerful foundation models (e.g., TimeSformer~\cite{bertasius2021space}, VideoMAE~\cite{tong2022videomae}) trained on vast and diverse video datasets like Kinetics~\cite{carreira2017quo, caba2015activitynet, goyal2017something, grauman2022ego4d, grauman2024ego}. COMODO provides a principled way for the wearable computing community to directly tap into this wealth of knowledge, transferring the nuanced understanding of human actions from the visual domain to the resource-constrained IMU domain, all without requiring a single IMU label. Our strong cross-dataset generalization results (Table~\ref{tab:transferability}) are crucial evidence for this; they indicate that the distilled knowledge captures transferable semantic structure of human activities, rather than superficial, dataset-specific correlations.

Second, COMODO is strategically positioned for the next generation of wearable devices. As smart glasses and other multimodal wearables become more prevalent, the collection of naturally occurring, temporally aligned video and IMU data will become common. In this emerging ecosystem, COMODO offers a way to bypass the need for manual IMU annotation entirely. Instead of facing the daunting and often impractical task of labeling hours of abstract IMU signals, developers can directly leverage these paired, unlabeled data streams, training a high-performance, deployment-ready IMU model directly. Furthermore, by breaking the dependency on IMU datasets with limited activity classes~\cite{zhang2012usc, anguita2013public, reiss2012introducing, chavarriaga2013opportunity}, COMODO opens the door to recognizing a much wider, more personalized "long tail" of human activities. Current HAR systems are often limited to a dozen or so generic activities like walking or running. However, real human life is composed of a vast array of nuanced, context-specific activities that span, but are not limited to domestic routines, occupational tasks, creative and leisure pursuits, and social interactions (see Section~\ref{sec:labels} for activity categories across datasets). It is infeasible to collect labeled IMU data for all such activities. By leveraging video, which is easier to find for niche activities, our approach provides a scalable path towards creating systems that understand the unique facets of an individual's life, a promising direction of proactive health and wellbeing applications.

\subsection{Implications for Ubiquitous and Wearable Systems: Designing for Practicality}
The design of ubiquitous and wearable systems is a constant exercise in balancing computational power, user experience, and practical constraints. COMODO contributes directly to this by addressing the critical trade-off between sensing fidelity and deployment feasibility.

First, our framework provides a concrete methodology for decoupling training-time requirements from deployment-time constraints. Continuous wearable video is often untenable due to power consumption and privacy concerns~\cite{10.1145/3613904.3642242, 10.1145/3613904.3642164, 10.1145/3706598.3713391, 10.1145/3432700}. COMODO addresses these barriers by learning from video offline and operating with IMU online, reconciling semantic richness with the requirements of unobtrusive, human-centered sensing. It allows systems to benefit from the wisdom of the "powerful but impractical" modality (video) while deploying the "practical but data-poor" modality (IMU).

Second, the model-agnostic nature of our framework (demonstrated in Table~\ref{tab:diff_teacher_student}) offers significant flexibility for system designers. It suggests that the COMODO paradigm is not tied to a specific architecture but is a general principle. As more powerful video or time-series foundation models emerge, they can be readily integrated as improved teachers or students, allowing for the continuous evolution of wearable intelligence without redesigning the entire system from the ground up.

\subsection{Limitation and Future Work}

\revision{Despite these advantages, certain constraints remain. First, COMODO relies on the availability of synchronized video-IMU data pairs during training. While this dependency currently limits the use of standalone IMU datasets, the growing adoption of next-generation wearable devices (e.g., smart glasses) is expected to facilitate the collection of naturally occurring, temporally aligned video–IMU data. Second, precise temporal alignment between heterogeneous modalities remains a challenge. Nevertheless, our similarity-distribution alignment inherently mitigates this issue by design; by preserving the teacher model's semantic structure, COMODO ensures that semantically similar segments are not equally penalized as negatives. Lastly, COMODO's effectiveness is inherently tied to the representational capacity of the pretrained video teacher, although its model-agnostic design allows it to benefit from the continuous advancement of video foundation models. Future research could focus on enhancing robustness against more severe asynchronicity to further reduce data requirements.}

While our current framework employs a one-way knowledge transfer from a pretrained, frozen teacher, future work could investigate mutual distillation frameworks, where the video and IMU models learn collaboratively, potentially allowing the teacher to also refine its representations based on the fine-grained temporal patterns captured by the IMU.  Additionally, our current setup does not explicitly consider personalization or contextual adaptation; integrating continual learning~\cite{parisi2019continual,10.1145/3530910} and user-specific adaptation mechanisms could make the framework more responsive to individual behavioral nuances. Finally, the distillation paradigm proposed in COMODO is not limited to HAR. We believe this approach of leveraging a data-rich modality to supervise a data-scarce modality may serve as a useful paradigm for other ubiquitous sensing tasks, such as transferring knowledge from video to audio for acoustic scene analysis or from video to physiological sensors for affective computing. We hope this perspective can inspire future work toward more general and flexible multimodal learning frameworks for ubiquitous and human-centered AI.
\section{Conclusion}
In this work, we introduced COMODO, a cross-modal self-supervised distillation framework designed to address a fundamental trade-off in human-centered computing: the interplay between the rich semantics of video and the practical deployability of IMU sensors. By distilling knowledge from a powerful, frozen video encoder to a lightweight IMU encoder via a dynamic instance queue, COMODO effectively bypasses the need for costly manual annotation of sensor data. Our extensive experiments demonstrate that this approach not only matches or surpasses supervised baselines but, more importantly, learns representations with strong cross-dataset generalization, a critical requirement for real-world utility. COMODO's model-agnostic and flexible design offers a principled and scalable path forward, enabling future wearable systems to leverage the immense progress in visual foundation models. We believe this work represents a significant step towards creating more capable, efficient, and privacy-preserving wearable AI.

% In this work, we introduced COMODO, a cross-modal self-supervised distillation framework that leverages pretrained video representations to enhance IMU-based egocentric HAR. By distilling rich semantic knowledge from a frozen video encoder to an IMU encoder through a dynamic instance queue, COMODO effectively mitigates the challenges posed by limited sensor data and costly annotations. Extensive experiments across multiple egocentric HAR benchmarks demonstrate that COMODO achieves results comparable to or even surpassing fully supervised baselines. Notably, our method exhibits strong cross-dataset generalization, underscoring its robustness and practical utility. COMODO's adaptability to various pretrained video and sensor models offers a viable framework for further investigating cross-modal knowledge transfer from visual models to resource-constrained sensors, enhancing future research in visually-guided, sensor-based HAR.

\begin{acks}
This research includes computations using the Wolfpack computational cluster, supported by the School of Computer Science and Engineering at UNSW Sydney. We also acknowledge support from the ARC Centre of Excellence for Automated Decision-Making and Society (CE200100005).
\end{acks}

\bibliographystyle{ACM-Reference-Format}
\bibliography{sample-base}

\appendix

\clearpage

\section{Baseline Implementation Detail}
\label{appendix:baselines}

To ensure a fair and rigorous evaluation, we meticulously configured our supervised and self-supervised baselines.

For the \emph{supervised baselines}, we include state-of-the-art time-series (DLinear ~\cite{zeng2023transformers}, Informer ~\cite{zhou2021informer}, TimesNet ~\cite{wu2023timesnet}) and HAR-specific (DeepConvLSTM~\cite{s16010115}, Attend and Discriminate~\cite{abedin2021attend}) models, alongside the same foundation models used as our student (MOMENT-small ~\cite{10.5555/3692070.3692712}, and Mantis ~\cite{feofanov2025mantis}). \revision{Additionally, we include recent state-of-the-art HAR frameworks: IMUGPT 2.0~\cite{leng2024imugpt} (in both \textit{Virtual-only} and \textit{Virtual+Real} settings) and the finetuned version of CrossHAR~\cite{hong2024crosshar}.} We use the official implementations of DLinear, Informer, and TimesNet from the Time-Series-Library\footnote{\url{https://github.com/thuml/Time Series Library}}, while DeepConvLSTM, Attend and Discriminate, MOMENT-small, Mantis, \revision{CrossHAR, IMUGPT 2.0} are implemented using their respective official repositories. Since both Ego4D and EgoExo4D have a standardized IMU sequence length of 1000 after preprocessing, we set the maximum sequence length to 1000 for these datasets. However, for MMEA, which has short and varying sequence lengths. Handling this variability is crucial for achieving optimal performance for each model. Therefore, we conducted a systematic evaluation on MMEA to determine the strongest configuration for different baseline categories. We explore three different settings: 
(1) The maximum sequence length of the entire dataset, which is 931.
(2) A fixed sequence length of 512, aligning with the input constraints of MOMENT-small and Mantis.
(3) A fixed sequence length of 100 and its augmented version.

Notably, Foundation models like MOMENT-small and Mantis are inherently capable of handling variable-length sequences. MOMENT-small leverages attention masking with left padding to process sequences of different lengths, while Mantis employs a built-in resize function to internally standardize all inputs to a fixed length (e.g., 512). These native capabilities allow them to adapt to the MMEA dataset's characteristics without requiring external segmentation.

\begin{table}[htbp]
\centering
\small
\begin{tabular}{ccccc}
\toprule
\textbf{Model} & \textbf{Max Seq Len} & \textbf{A@1} & \textbf{A@3} & \textbf{A@5} \\
\midrule
\multirow{3}{*}{DLinear} 
 & 100 & 19.36 & 33.99 & 44.44 \\
 & 512 & \underline{22.79} & \underline{37.73} & \underline{45.81} \\
 & 931 & \textbf{24.01} & \textbf{37.73} & \textbf{47.18} \\
\midrule
\multirow{3}{*}{Informer} 
 & 100 & 37.12 & 63.19 & 75.46 \\
 & 512 & \textbf{47.10} & \textbf{70.66} & \textbf{81.10} \\
 & 931 & \underline{45.81} & \underline{67.68} & \underline{78.81} \\
\midrule
\multirow{3}{*}{TimesNet} 
 & 100 & 41.92 & 67.23 & 77.82 \\
 & 512 & \underline{48.48} & \textbf{74.47} & \underline{83.23} \\
 & 931 & \textbf{50.30} & \underline{74.24} & \textbf{83.99} \\
\midrule
\multirow{4}{*}{DeepConvLSTM} 
 & 100\(^{\dag}\) & \textbf{52.58} & \textbf{77.25} & \textbf{86.44} \\
 & 100 & \underline{38.68} & \underline{64.21} & \underline{76.75} \\
 & 512 & 12.54 & 29.26 & 42.55 \\
 & 931 & 3.72 & 10.64 & 17.10 \\
 \midrule
\multirow{4}{*}{Attend and Discriminate} 
 & 100\(^{\dag}\) & \underline{58.66} & 79.87 & 87.47 \\
 & 100 & 47.72 & 72.19 & 82.07 \\
 & 512 & \textbf{65.43} & \textbf{87.31} & \textbf{93.24} \\
 & 931 & 58.66 & \underline{83.21} & \underline{91.03} \\
 \midrule
 \revisionrow
 \multirow{3}{*}{CrossHAR}
 & 100 & \underline{64.29} & \underline{86.48} & \underline{92.50} \\
 & \revisionrow 512 & 8\textbf{4.97} & \textbf{96.55} & \textbf{98.11} \\
 \revisionrow
 & 931 & OOM & OOM & OOM \\
\bottomrule
\end{tabular}
\caption{Comparison of the impact of different max sequence length for supervised baselines on MMEA dataset. \({\dag}\) denotes models trained with non-overlapping fixed-length segmentation of IMU sequences. The best are indicated in \textbf{bold}, and the second-best are \underline{underlined}.}
\label{tab:tslib}
\end{table}

Table \ref{tab:tslib} presents the performance of these models under different sequence length settings on the MMEA dataset. Based on these results, we select specific configurations for the comparison in the main result of our paper. For time-series models (DLinear, Informer, and TimesNet), we observe that DLinear and TimesNet achieve better accuracy when using the full sequence length of 931, while Informer performs slightly better with a sequence length of 512. To maintain consistency in baseline configurations, we report the results using 931 in the main paper, providing a strong and representative baseline. 

For the HAR-specific models, DeepConvLSTM and Attend and Discriminate, we report results under the non-overlapping fixed-length segmentation setting (100\(^\dag\)). This choice ensures configuration consistency across HAR models, as DeepConvLSTM performs poorly with longer sequences due to its sensitivity to zero-padding and the lack of mechanisms (such as attention or masking) to mitigate the influence of padded inputs. By segmenting longer sequences into fixed chunks of length 100, we not only avoid padding but also augment the dataset (by approximately 3\(\times\)), thereby enabling DeepConvLSTM to achieve its best performance. While Attend and Discriminate achieves its highest accuracy at a sequence length of 512, it also performs competitively under 100\(^\dag\). To maintain consistency across HAR baselines, we adopt the 100\(^\dag\) setting for both models in our main comparison. \revision{Regarding CrossHAR, despite its 20 Hz downsampling, its Transformer-based architecture and large batch size requirements lead to prohibitive memory usage. It encounters Out-of-Memory (OOM) errors on an NVIDIA L40S GPU (48GB) at a sequence length of 931; therefore, we report its best-performing stable configuration at a sequence length of 512. For IMUGPT 2.0, we follow to their original protocol which employs a built-in sliding window mechanism (2-second duration with 50\% overlap), requiring no additional segmentation processing.}

Crucially, our proposed COMODO framework substantially outperforms all baseline models, regardless of their configuration in Table \ref{tab:tslib}. Therefore, the specific choice of baseline settings does not affect the overall conclusion about the superiority of our method.

Our \emph{self-supervised baselines} include two pretrained time-series foundation models, MOMENT-small~\cite{10.5555/3692070.3692712} and Mantis~\cite{feofanov2025mantis}, as well as \revision{CrossHAR~\cite{hong2024crosshar}, a hierarchical self-supervised framework for IMU-based HAR, and} IMU2CLIP~\cite{moon2023imu2clip}, a contrastive cross-modal framework. For all self-supervised methods, we follow the standard evaluation protocol in representation learning to assess the quality of their learned representations (as detailed in Section~\ref{sec:inference}). This involves two stages: first, we use the pretrained encoder to extract feature representations from the data; second, we freeze the encoder and train a separate classifier on these features. This protocol allows us to isolate and measure the discriminative power of the representations themselves, independent of the final classification head.

Given that IMU2CLIP is the most direct methodological competitor, ensuring a fair comparison is important. Therefore, we undertook a careful reimplementation of IMU2CLIP while preserving its core cross-modal contrastive learning framework. In the original IMU2CLIP, an IMU encoder is trained to align IMU representations with CLIP-based text and video embeddings using the InfoNCE loss~\cite{moon2023imu2clip}. However, several key differences exist between IMU2CLIP and our proposed approach, particularly in the choice of teacher and student networks. To enable a fair baseline comparison, we introduced the following modifications:

(1) Consistent Teacher and Student Architectures: IMU2CLIP originally employs CLIP-based image embeddings as the video representation. In our reimplementation, we replace this with our video teacher model to ensure a consistent teacher architecture across baselines. Similarly, instead of IMU2CLIP’s custom IMU encoder, we use our selected IMU models (Mantis or MOMENT-small).

(2) Feature Dimension Matching: IMU2CLIP originally maps IMU representations into a CLIP-aligned embedding space, where the feature dimensions of CLIP and IMU are naturally the same, eliminating the need for additional transformations. However, in our reimplementation, the video models and IMU student models have mismatched feature dimensions. To address this, we introduce a linear projection layer (without activation) to align the IMU embeddings with the video model’s feature space.

(3) Preserving Contrastive Training: IMU2CLIP uses the InfoNCE loss to encourage alignment between IMU and video embeddings. We retain this objective, ensuring that the primary distinction between IMU2CLIP and COMODO lies in the distillation mechanism rather than in major architectural differences.

We also ensure that the IMU2CLIP reimplementation follows the same temporal alignment strategy, where each IMU segment corresponds to the same time span as the video frames. Additionally, we apply the same data augmentation techniques as COMODO and adopt the same hyperparameters (learning rate, epochs, and batch size) to ensure a fair experimental comparison.

In addition to architectural modifications, there exist discrepancies in dataset availability that impact the reproducibility of IMU2CLIP’s results. Specifically, while IMU2CLIP reports results on Ego4D, the IMU2CLIP paper does not provide official benchmark splits. To address this, we reprocessed the available Ego4D (v2) data while maintaining a 7:2:1 train-test-validation split to ensure a consistent comparison across methods. As part of this process, we removed activity classes with fewer than 10 samples to enable meaningful evaluation. Our reimplementation adheres to IMU2CLIP's contrastive learning framework while aligning with the teacher-student configuration used in COMODO, enabling an equitable performance evaluation. Although we use a single GPU, we match the effective global batch size (16×8 GPUs) in IMU2CLIP by setting the batch size to 128. This results in a negative sample pool size of 127, the same as the original.

\end{document}